\documentclass[10pt,twocolumn,letterpaper]{article}

\usepackage[pagenumbers]{cvpr} 
\usepackage[accsupp]{axessibility} 

\usepackage[dvipsnames]{xcolor}
\usepackage{makecell}
\usepackage{multirow}
\usepackage{rotating}
\usepackage[hang,flushmargin]{footmisc}

\newcommand{\suppmat}[1]{\color{black}#1}
\usepackage[dvipsnames]{xcolor}
\newcommand*\samethanks[1][\value{footnote}]{\footnotemark[#1]}

\newcommand{\mytilde}{\raise.17ex\hbox{$\scriptstyle\mathtt{\sim}$}}

\newcommand{\Epsilon}{\mathcal{E}}
\newcommand{\modelname}{MatFuse\xspace}

\definecolor{cvprblue}{rgb}{0.21,0.49,0.74}
\usepackage[pagebackref,breaklinks,colorlinks,citecolor=cvprblue]{hyperref}
\usepackage{adjustbox}
\def\paperID{5499} 
\def\confName{CVPR}
\def\confYear{2024}

\title{\modelname: Controllable Material Generation with Diffusion Models}

\author{Giuseppe Vecchio\thanks{Both authors contributed equally to this research.}
\qquad
Renato Sortino\samethanks[1]
\qquad
Simone Palazzo
\qquad
Concetto Spampinato\\
{\tt\small giuseppe.vecchio@phd.unict.it \qquad renato.sortino@phd.unict.it} \\
University of Catania\\
}

\let\oldtwocolumn\twocolumn
\renewcommand\twocolumn[1][]{%
    \oldtwocolumn[{#1}{
    \begin{center}
            \captionsetup{type=figure}
            \includegraphics[width=\textwidth]{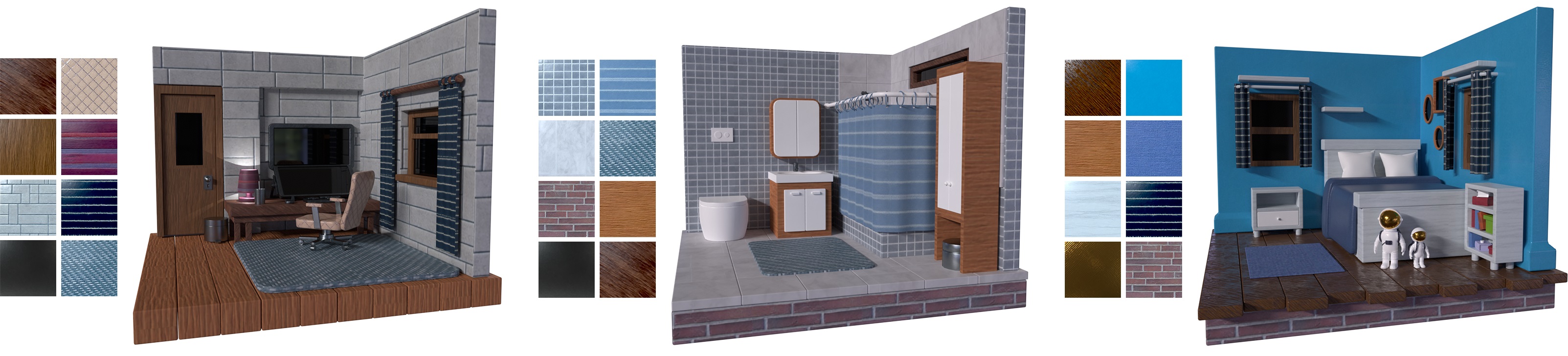}
            \captionof{figure}{\textbf{Sample scenes textured using materials generated with MatFuse.} For each of the three scenes we show the materials used and the final rendering.}
            \label{fig:thumbnail}
        \end{center}
    }]
}

\begin{document}
\maketitle
\begin{abstract}
Creating high-quality materials in computer graphics is a challenging and time-consuming task, which requires great expertise. 
To simplify this process, we introduce \textbf{\modelname}, a unified approach that harnesses the generative power of diffusion models for creation and editing of 3D materials.
Our method integrates multiple sources of conditioning, including color palettes, sketches, text, and pictures, enhancing creative possibilities and granting fine-grained control over material synthesis.
Additionally, \modelname enables map-level material editing capabilities through latent manipulation by means of a multi-encoder compression model which learns a disentangled latent representation for each map.
We demonstrate the effectiveness of \modelname under multiple conditioning settings and explore the potential of material editing. Finally, we assess the quality of the generated materials both quantitatively in terms of CLIP-IQA and FID scores and qualitatively by conducting a user study. \\
Source code for training \modelname and supplemental materials are publicly available at {\tt\small \url{https://gvecchio.com/matfuse}}.
\end{abstract}    
\section{Introduction}
Materials are central in computer graphics, playing a pivotal role in achieving high-quality, realistic digital imagery. As the computational power of professional and consumer hardware has increased, high-quality CGI has experienced a growing demand, fueled by the expanding field of application of 3D models, from game engines to architectural and industrial prototyping and simulation~\cite{vecchio2022midgard, song2023synthetic, infinigen2023infinite}.
However, the creation of high-quality materials remains a challenging and time-consuming process, which requires complex tools and high expertise.
Following the promising results achieved by Generative Adversarial Networks (GANs)~\cite{goodfellow2014generative} for the generation of natural images, several works have successfully employed adversarial training to generate high-quality materials~\cite{guo2020materialgan,zhou2022tilegen,hu2022controlling,guo2023text2mat}. These approaches, however, provide a limited degree of control over material synthesis. Additionally, GANs are generally hard to train, due to the inherent instability of their adversarial training, leading to mode collapse and limited variability.
 
Recently, diffusion models (DMs) have set a new state-of-the-art in image generation~\cite{ho2020denoising,dhariwal2021diffusion,rombach2022high}, overcoming the training limitations of GANs. 
Furthermore, diffusion models can be easily conditioned during the ``denoising'' process, with \textit{global} or \textit{local} conditions, respectively controlling the overall appearance of the image (e.g., text prompt), or specific regions of the output image (e.g., sketches). 
Recent approaches like Composer~\cite{huang2023composer}, propose to combine multiple sources of conditioning, both global and local, by considering an image as the sum of its independent components~\cite{lake2017building}. This approach expands the control space, giving designers the degree of control required to finely guide the generation. 

We design \textbf{\modelname} to improve material synthesis using the generative capabilities of diffusion models and exploiting the image compositionality approach to combine multiple conditioning sources in a single model.
Following \citet{rombach2022high}, the proposed model consists of a VQ-GAN~\cite{esser2021taming}, trained to learn a bidirectional mapping between the pixel space and the latent space, and a diffusion model, trained to generate a latent representation of a material starting from noise and one or more optional conditions. 

We evaluate the effectiveness of our approach when being conditioned with both a single or multiple conditions. We also test our model for material editing through what we define as \textit{volumetric inpainting}, by partially or totally masking, single SVBRDF maps for a given material, and letting the model reconstruct the missing parts.
The results show the potential of \modelname in generating a wide range of diverse and realistic materials, as well as in adapting to several combinations of conditioning inputs. \\
In summary, the contributions of this work are: 
\begin{itemize}
    \item We present \modelname, a unified, multi-conditional method leveraging the generation capabilities of diffusion models to tackle the task of high-quality material synthesis as a set of SVBRDF maps.
    \item We propose a multi-encoder extension to the auto-encoder by~\citet{rombach2022high}, using 4 different encoders, to learn map-specific latent spaces and add a rendering loss to its training.
    \item We demonstrate the generation capabilities and flexibility of \modelname, through different conditioning mechanisms, which allow for an unprecedented level of control for material generation.
    \item We show the ability to use \modelname for material editing purposes through ``volumetric inpainting'', to generate single portions of input materials or entire maps.
\end{itemize}
\section{Related Work}
\label{sec:related}

\begin{figure*}[ht]
    \centering
    \includegraphics[width=.8\linewidth]{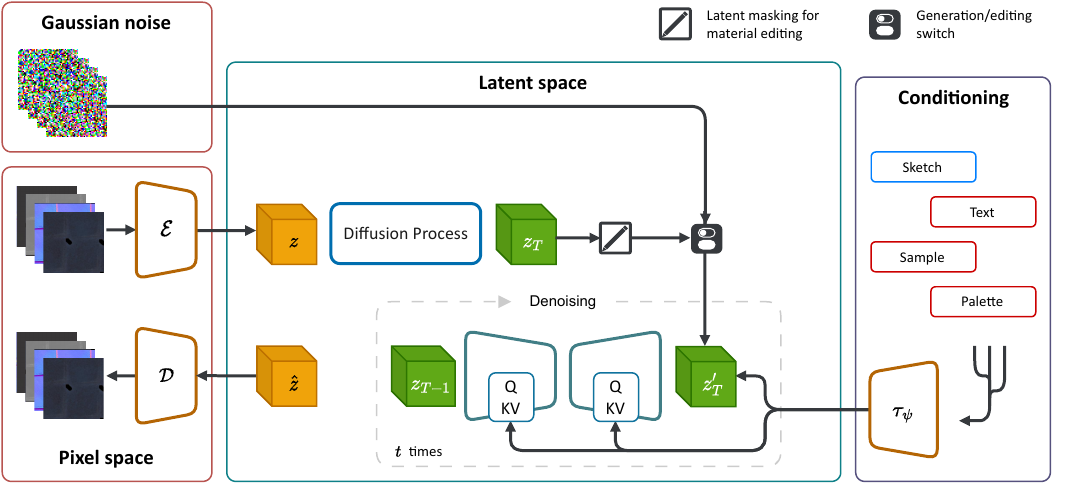}
    \caption{\textbf{Overview of the \modelname framework}: At training time, VQ-GAN encoder $\mathcal{E}$ projects data from the pixel space to a more compact latent embedding $z$; the diffusion process runs on this latent space; conditioning is carried out through cross-attention for global conditions (red in figure), and through concatenation with the noise for local conditions (blue in figure); the output maps are finally obtained by projecting the conditioned reconstructed latent space $\hat{z}$ back into the pixel space through VQ-GAN decoder $\mathcal{D}$.}
    \label{fig:overview}
\end{figure*}

\noindent\textbf{Controllable material generation.}
Materials synthesis is a challenging task in computer graphics~\cite{guarnera2016brdf}, with many recent data-driven approaches focusing on the task of estimating SVBRDF maps from an input image~\cite{deschaintre2018single,li2017modeling,li2018materials,martin2022materia,guo2021highlight,zhou2021adversarial,vecchio2021surfacenet,aittala2016reflectance,deschaintre2019flexible,bi2020deep,gao2019deep}.

Controllable generation of materials, in contrast, remains a relatively underexplored task. 
\citet{guehl2020semi} propose an approach consisting of a procedural structure synthesis step, followed by data-driven color synthesis to propagate existing material properties to the generated structure.
MaterialGAN~\cite{guo2020materialgan} proposes a generative network based on StyleGAN2~\cite{stylegan2}, trained to synthesize realistic SVBRDF parameter maps. This approach exploits the properties of the latent space learned by StlyleGAN2 to generate material maps that match the appearance of the captured images when rendered.
\citet{hu2022controlling} extend the capabilities of MaterialGAN with the generation of novel materials, by transferring the micro- and meso-structure of a texture to a set of input material maps. However, both MaterialGAN~\cite{guo2020materialgan} and \citet{hu2022controlling} rely on alterations of pre-existing material inputs and lack any generation capabilities.
Recently, \citet{zhou2022tilegen} proposed TileGen, a generative model for SVBRDFs capable of producing tileable materials, optionally conditioned through an input structure pattern. However, its generation capabilities are strongly limited to class-specific training. 
\citet{guo2023text2mat} proposed Text2Mat, an architecture based on diffusion models for text-to-material generation. 
However, controlling the generation of materials remains a challenging task. To fill this gap, \modelname leverages diffusion models to provide full control and flexibility over the generation process by ingesting multiple conditions to guide the diffusion process.

\noindent\textbf{Generative models.} Image generation is a long-standing challenge in computer vision due to the high dimensionality of images and the difficulty in modeling complex data distributions. Generative Adversarial Networks (GAN)~\cite{goodfellow2014generative} enabled the generation of high-quality images~\cite{karras2017progressive,brock2018large,karras2020analyzing} but are characterized by unstable convergence at training time~\cite{arjovsky2017wasserstein,gulrajani2017improved,mescheder2018convergence}, due to the adversarial training, and are unable to fully model complex data distributions~\cite{metz2016unrolled}, often exhibiting mode collapse behavior.

Recently, Diffusion Models (DMs)~\cite{sohl2015deep,ho2020denoising} have emerged as an alternative to GANs, achieving state-of-the-art results in image generation tasks~\cite{dhariwal2021diffusion}, besides showing a more stable training behavior.
However, optimizing these models tends to be expensive in terms of training times and computational costs. To address these limitations, \citet{rombach2022high} propose to apply the diffusion process to a smaller, and less computationally demanding, latent space, perceptually equivalent to the pixel space.
This shift to the latent space reduces computational requirements, without altering generation quality, and enables a whole new classifier-free conditioning mechanism~\cite{ho2022classifier} through cross-attention between latent image representations and conditioning data.
More recently, Composer~\cite{huang2023composer} showed how it is possible to combine multiple semantically different conditions to control diffusion models. 

Building on these advancements, \modelname a) introduces a multi-encoder VQ-GAN to account for individual SVBRDF map peculiarities and integrates a rendering loss~\cite{deschaintre2018single} to enforce coherence and consistency in the output results; b) extends the LDM conditioning mechanism to include multiple modalities in a compositional way; c) enables inpainting both at spatial and map level, by exploiting the multi-encoder approach, providing users fine control over the generation process. 
\section{MatFuse Architecture}
\label{sec:method}

Motivated by the lack of a unified model capable of accepting different sources of control for material generation, we propose \modelname, a conditional generative model that produces high-quality pixel-level reflectance properties for arbitrary materials, while simultaneously combining multiple conditions. 
To this end, we leverage the compositionality of images, by deconstructing them into primitives, such as color palettes, sketches, etc. which can then be combined to guide the generation~\cite{huang2023composer}.

Inspired by the LDM~\cite{rombach2022high} architecture, \modelname consists of two main components: 1) a compression network that projects data from the pixel space $\mathcal{X}$ to the latent space $\mathcal{Z}$ and vice-versa, and 2) a diffusion model, which learns the distribution of the latent feature vectors to enable the generation of new samples.

The general architecture of \modelname is shown in Fig.~\ref{fig:overview}.\\
In the following, we introduce and describe each module of the proposed framework.

\subsection{Latent Diffusion Model}
\label{sec:compression}

\begin{figure}
    \centering
    \includegraphics[width=1\linewidth]{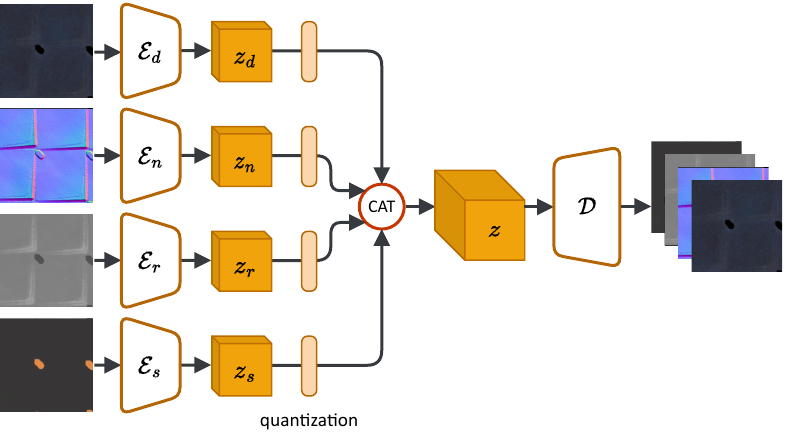}
    \caption{\textbf{Overview of the compression model architecture}. Reflectance maps (diffuse, normal, roughness, and specular) are fed to the encoders. Features extracted for each map are quantized and concatenated before being passed to the decoder, which reconstructs the original maps.}
    \label{fig:vq_gan}
\end{figure}

\noindent \textbf{Map Compression}.
We use a multi-encoder VQ-GAN~\cite{esser2021taming} to learn a map-specific latent representation.
This allows the model to extract disentangled features from each map, which will be concatenated in the latent space and combined by the diffusion model via self-attention. The architecture is illustrated in Fig.~\ref{fig:vq_gan}.

Given a set of $N$ maps $x = \left\{ \textbf{M}^1, \textbf{M}^2, \dots, \textbf{M}^N\right\}$ and encoders $\Epsilon = \left\{ \mathcal{E}^1,  \mathcal{E}^2, \dots, \mathcal{E}^N \right\}$, each map $\textbf{M}^i \in \mathbb{R}^{H \times W \times 3}$ is encoded into a latent representation $z^i = \mathcal{E}^i(\textbf{M}^i)$, where $z^i \in \mathbb{R}^{h \times w \times c^i}$, $i \in \left\{1, \dots, N\right\}$, and $c^i$ is the number of channels of each encoded map.
The $N$ latent representations (one per input map) are then concatenated along the channel dimension, obtaining $z = concat\left(z^1, z^2, \dots, z^N\right)$, and then fed into the decoder $\mathcal{D}$ that reconstructs the set of input maps $\hat{x} = \left\{ \hat{\textbf{M}}^1, \hat{\textbf{M}}^2, \dots, \hat{\textbf{M}}^N\right\}$. Here, $\hat{x} = \mathcal{D}(z)$, $ z \in \mathbb{R}^{h \times w \times c}$, where $c$ is the number of channels of the concatenated maps, i.e., $c = \sum_i{c^i}$. Before decoding the feature vectors, we regularize the latent space by learning a representative \emph{codebook} for each map, which is then used to quantize the latent vectors $z_i$ before concatenation.

Following the work of~\citet{rombach2022high}, we train the encoder $\Epsilon$ using a combination of pixel-space $L_2$ loss $\mathcal{L}_\text{pixel}$, a perceptual LPIPS loss $\mathcal{L}_\text{perc}$~\cite{zhang2021designing}, a patch-based adversarial objective $\mathcal{L}_\text{adv}$~\cite{isola2017image, dosovitskiy2016generating, esser2021taming}, and a codebook commitment loss $\mathcal{L}_\text{comm}$~\cite{van2017neural}.
To improve the reconstruction of material map details in the latent space, we add a rendering loss $\mathcal{L}_\text{render}$~\cite{deschaintre2018single}, computed as the MSE between ground-truth renders and prediction renders, to the VQ-GAN training, enforcing coherence and consistency between the individual maps.

\noindent \textbf{Diffusion Model}.
After learning a latent space that efficiently encodes information from multiple maps, we train a diffusion model~\cite{ho2020denoising} to estimate the prior distribution of the latent vectors to synthesize real samples. We follow the architecture proposed in~\citet{rombach2022high}, which consists of a U-Net~\cite{ronneberger2015u} with self-attention between residual blocks operating on the latent representation $z$ of the input maps, rather than on the pixel space.

In particular, the diffusion network $\epsilon_{\theta}$ is trained to estimate at each step the noise added in the forward diffusion process and subtracts it from the noisy latent to obtain the denoised data.
We optimize the diffusion model with an $\mathcal{L_\text{diff}}$ objective between the estimated noise and the noise added in the forward diffusion process, as in \citet{ho2020denoising}.
\begin{equation}
    \mathcal{L_\text{diff}}=\mathbb{E}_{t, z_0, \epsilon}\left[\left\|\epsilon_{t}-\epsilon_\theta\left(z_t, t\right)\right\|^2\right]
\end{equation}
Here, $\epsilon_{t}$ is the noise added at the timestep $t$ in the forward diffusion process, while $\epsilon_\theta\left(z_t, t\right)$ is the noise estimated by the U-Net model at time $t$.

The trained model allows for generating new samples by denoising the noise vector sampled from a normal distribution into a valid latent space point.

\subsection{Conditioning Mechanisms}
\label{sec:conditioning}

\modelname allows for controlling the generation process via two types of conditioning information: 1) global conditioning for a high-level control via text or visual prompts, as well as color palettes, and; 2) local conditioning, for a fine-grained localized control, via sketches.

\noindent \textbf{Global conditioning}.
It enables control over the generation via high-level prompts, descriptive of the global material appearance. Conditions are embedded in a one-dimensional feature vector, which is provided to the diffusion model through multi-head cross-attention~\cite{vaswani2017attention} at each denoising step between the flattened noise tensor $z$ and the conditioning vector (i.e., QKV in Fig.~\ref{fig:overview}).

\modelname can be globally conditioned via text and image prompts, as well as color palettes.
Image and text embeddings are computed using a pre-trained CLIP~\cite{clip} model as a feature extractor. The color palette embedding is computed by counting color occurrences in an input image, clustering those within a certain CIE76 distance threshold, and selecting the top 5 most prevalent colors. Finally, these values are projected into a 1D vector through an MLP, which is optimized in conjunction with the diffusion model.

\noindent \textbf{Local conditioning}. 
Local conditions are used to achieve control over the generation structure. These conditions are first projected into a low-resolution representation to match the dimensionality of the latent vector $z$ using a small convolutional network which is trained jointly with the U-Net. The resulting embedding is concatenated to the noisy latent $z$.
We identify sketches as a relevant local condition for materials, giving control over the represented pattern. We extract sketches from the material render, under a diffuse light, using a Canny edge detector~\cite{canny1986computational}.

\noindent \textbf{Multimodal fusion}.
We enable multimodal composable generation in \modelname by extending the classifier-free guidance training strategy~\cite{ho2022classifier}. 
This strategy allows not only to combine different types of conditioning but also to generate quality output regardless of the number of conditions provided, thus allowing compositionality.
In particular, during training, we randomly drop each condition with a probability of 50\% and drop all conditions with a 10\% probability.

With conditioning, the training objective becomes 
\begin{equation}
    \mathcal{L_\text{diff}}=\mathbb{E}_{t, z_0, \epsilon}\left[\left\|\epsilon_{t}-\epsilon_\theta\left(z_t, t, \tau(y)\right)\right\|^2\right]
    \label{eq:ldm_objective}
\end{equation}

\subsection{Material Editing via Volumetric Inpainting}
\label{sec:editing}

The use of a multi-encoder architecture, as described in Sec.~\ref{sec:compression}, allows the model to learn a disentangled latent representation of each material map, hardly achievable using a single encoder, by encoding each map separately. This latent representation allows us to manipulate specific parts of the latent space, knowing which material property they encode, thus enabling an unprecedented level of material editing capabilities. 
We propose a novel ``volumetric inpainting''~\footnote{The name was chosen to highlight that masking occurs in both the spatial and channel (encoding material property information) dimensions independently.} approach by jointly masking portions of the noise tensor in both spatial and channel dimensions. 
Formally, given a latent representation $z$ of the material maps, encoded through $\mathcal{E}$, we compute the latent tensor at the denoising step $t$ as $z \odot m + z_{t+1} \odot (1 - m)$, with $m$ being the volumetric binary mask.

By masking portions or channels corresponding to a specific map, we can force the model to generate only that particular map. This is particularly useful for incomplete materials, missing some maps, or where the properties of one map are not satisfying. In combination with traditional inpainting, it is possible to generate only specific areas of the material for all maps or for a reduced set only. 

The application of ``volumetric inpainting'' for material editing is demonstrated in Sec.~\ref{sec:results_edit}.
\section{Experimental results}
\label{sec:results}

In this section, we first introduce the datasets employed in our work: the synthetic dataset by \citet{deschaintre2018single}, and a new procedurally-generated synthetic dataset created for the task at hand.
Then, we evaluate the accuracy of our approach on two different training setups: 1) with a single condition, providing examples for each condition individually, and 2) with multiple conditions combined together. We compare our approach against TileGen~\cite{zhou2022tilegen} and evaluate its performance in terms of CLIP-IQA~\cite{wang2023exploring} and Fréchet Inception Distance (FID)~\cite{heusel2017gans} to provide a quantitative measure. We also conduct a user study to evaluate the user preference when comparing the two methods.
In addition, we demonstrate the material editing capabilities of \modelname, thus confirming the advantage introduced by a multi-encoder architecture. 
Finally, we ablate the proposed architectural components, in particular the multi-encoder compression model, and losses to assess their contribution.

\subsection{Datasets}
\label{sec:datasets}
We employ the SVBRDF dataset introduced by~\cite{deschaintre2018single}, which is based on the Allegorithmic Substance Share collection\footnote{\url{https://share.substance3d.com/}}.
The entire dataset includes about 20,000 blended materials represented with the \emph{diffuse}, \emph{normal}, \emph{specular}, and \emph{roughness} maps. We use the training/test splits introduced by \citet{deschaintre2018single}.

We extended the dataset with 320 materials collected from the PolyHaven\footnote{\url{https://polyhaven.com/textures/}} library.
As a form of data augmentation for this dataset, we extract crops at different scales from each material at the 4K resolution. For each material, we collect 1 full-scale (4K resolution) crop, 4 crops at half-scale (2K resolution), 16 crops at a quarter of scale (1K resolution), 64 crops at one eight of resolution (512), and 256 crops at one-sixteenth of resolution (256). The full dataset consists of 431 crops for each material and a total number of 140,508 crops. Each crop is then rescaled to a resolution of 256$\times$256 pixels.\\
The combination of the two datasets sums up to about 160K materials.
We render each material under different lighting conditions using five different environment maps, each rotated four times (0°, 90°, 180°, 270°) to light the material from different angles. For each material crop, we produce 20 renders, for a total of 3.2 million renders.

\subsection{Training Procedure}
The compression model is optimized via a combination of supervised and adversarial training. It is trained with mini-batch gradient descent, using the Adam~\cite{adam} optimizer and a batch size of 4. The learning rate is set to $10^{-4}$ and the training is carried out for 4,000,000 iterations.
$\mathcal{L}_\text{adv}$ is enabled after 300,000 iterations, when the compression model starts to learn low-frequency components of maps, thus allowing the compression model to recover the high-frequency ones.
For our encoders we choose a downsampling factor of $f = 8$, giving the best compromise between efficiency and image quality, as shown in~\cite{rombach2022high}.

The diffusion model is trained for $500,000$ iterations with a batch size of $20$ using an AdamW~\cite{adamw} optimizer, with a learning rate value of $10^{-4}$, with a linear learning rate warm-up starting from $10^{-6}$. We used a linear schedule for $\beta$ and denoise using the DDIM (Denoising Diffusion Implicit Models)~\cite{song2020denoising} sampling schedule at inference time with $T=50$ steps.

\subsection{Generation Results}
\label{sec:results_gen}
We evaluate our model under different conditioning settings to understand its capability of combining multiple sources of information while producing high-quality material maps. In particular, we explore single-conditional and multi-conditional generation. \suppmat{Unconditional samples and additional conditional generation examples are included in the supplemental material.}

\subsubsection{Single-conditional generation}
We first assess the generation capabilities of \modelname when controlled with a single input condition.

\begin{figure}
    \centering
    \setlength{\tabcolsep}{.5pt}
    \begin{tabular}{cccccc}
        \hspace{-1mm}\small{Condition} & \small{Diffuse} & \small{Normal} & \small{Roughness} & \small{Specular} & \small{Render} \\
        
        \vspace{-1mm}\hspace{-1mm}\makecell[b]{\footnotesize{``terracotta} \\ \footnotesize{brick wall''}\vspace{2.5mm}} &
        \includegraphics[width=0.166\linewidth]{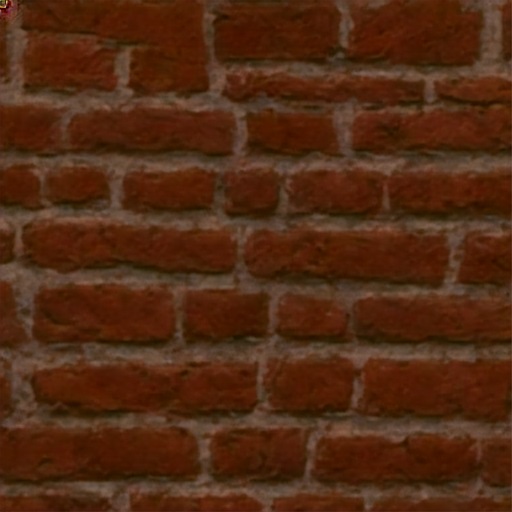} &
        \includegraphics[width=0.166\linewidth]{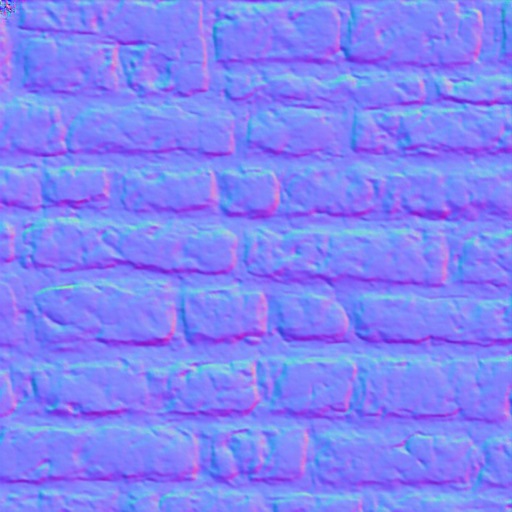} &
        \includegraphics[width=0.166\linewidth]{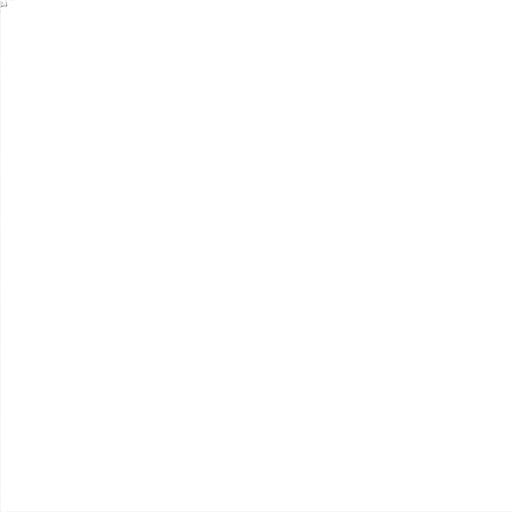} &
        \includegraphics[width=0.166\linewidth]{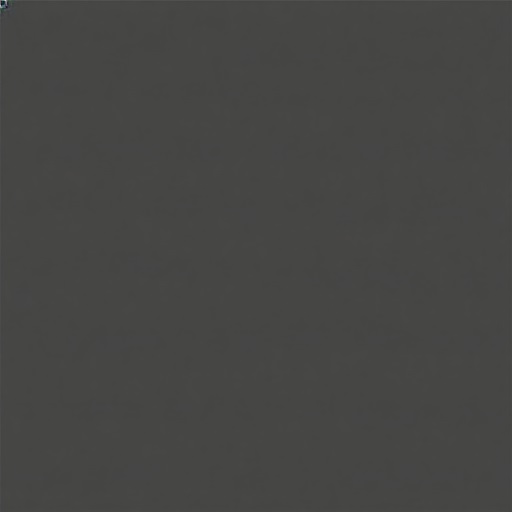} &
        \includegraphics[width=0.166\linewidth]{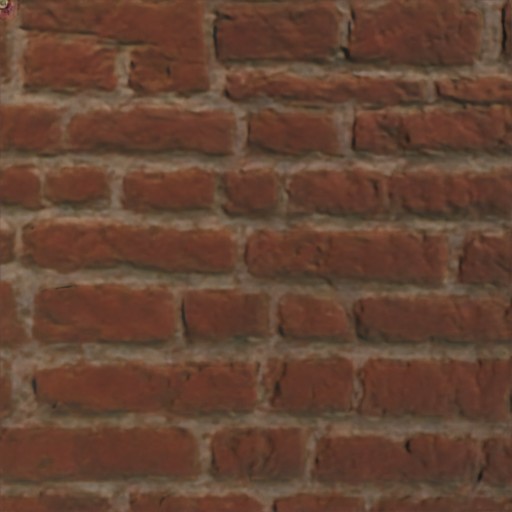} \\
    
        \vspace{-1mm}\hspace{-1mm}\makecell[b]{\footnotesize{``shiny} \\ \footnotesize{parquet''}\vspace{2.5mm}} &
        \includegraphics[width=0.166\linewidth]{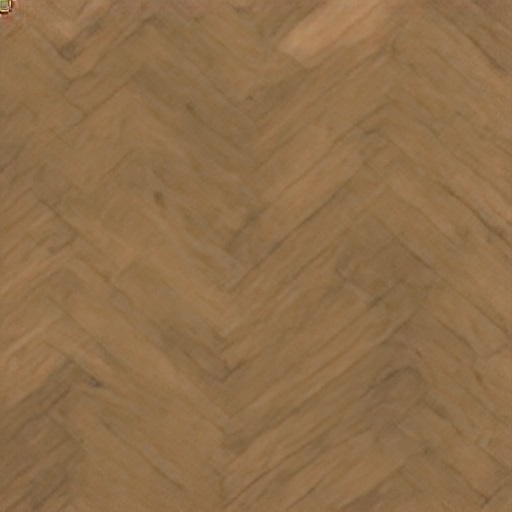} &
        \includegraphics[width=0.166\linewidth]{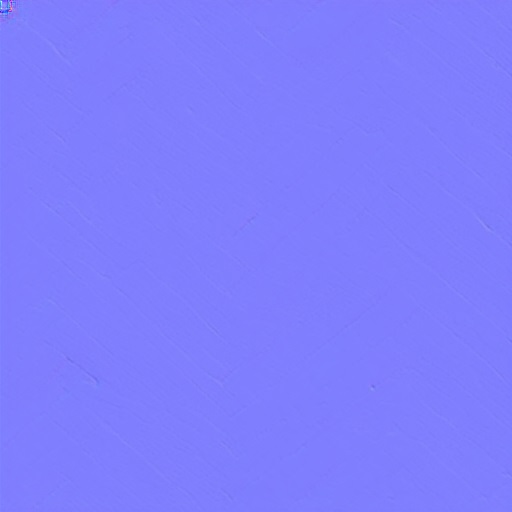} &
        \includegraphics[width=0.166\linewidth]{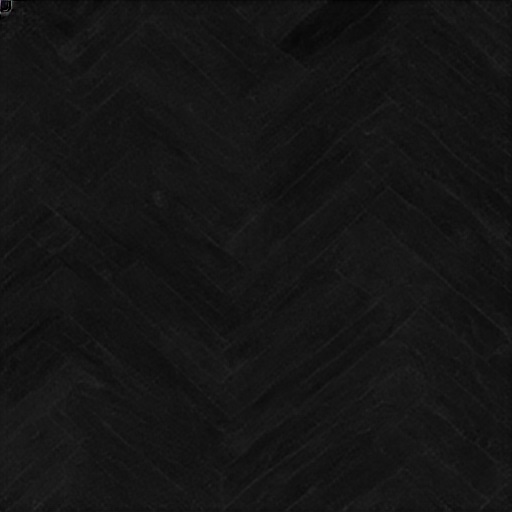} &
        \includegraphics[width=0.166\linewidth]{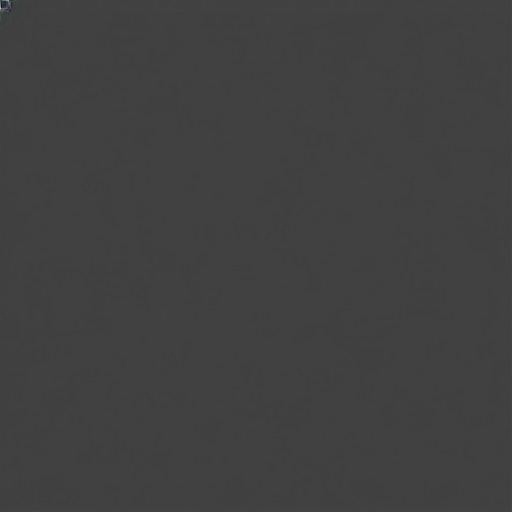} &
        \includegraphics[width=0.166\linewidth]{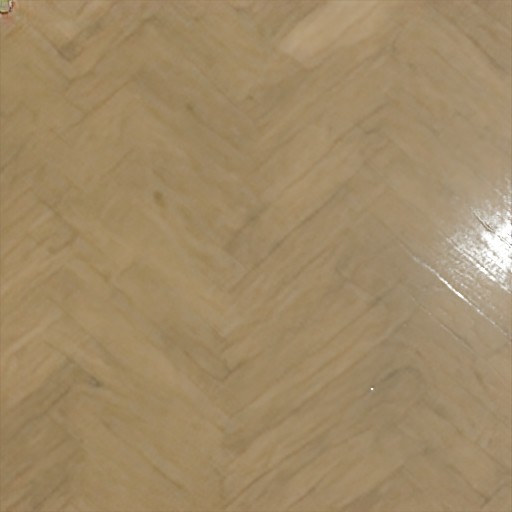} \\
    
        \vspace{-1mm}\hspace{-1mm}\includegraphics[width=0.166\linewidth]{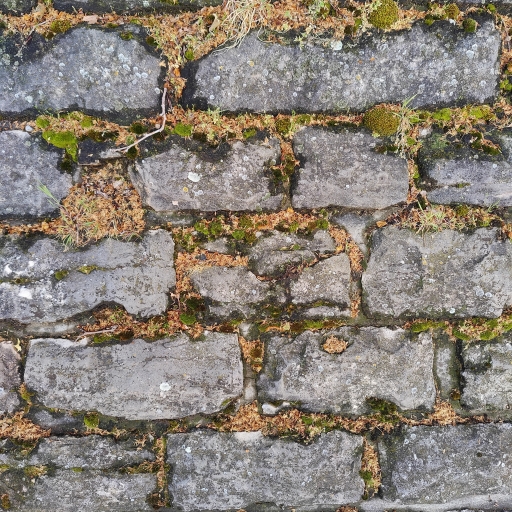} &
        \includegraphics[width=0.166\linewidth]{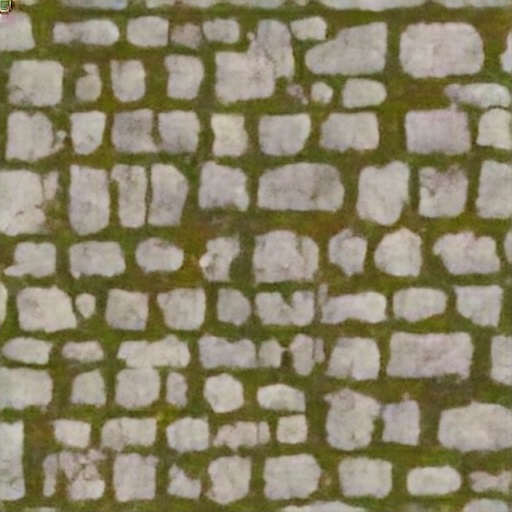} &
        \includegraphics[width=0.166\linewidth]{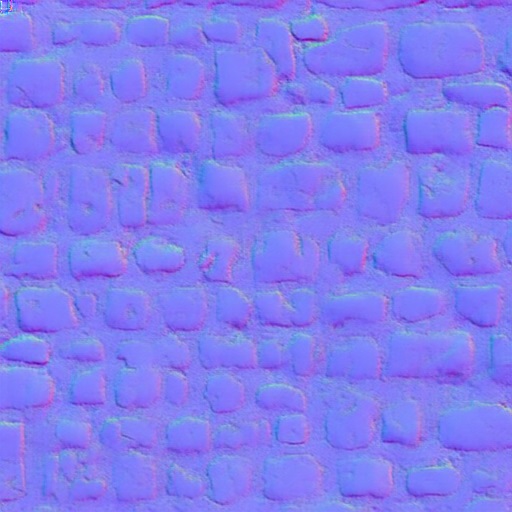} &
        \includegraphics[width=0.166\linewidth]{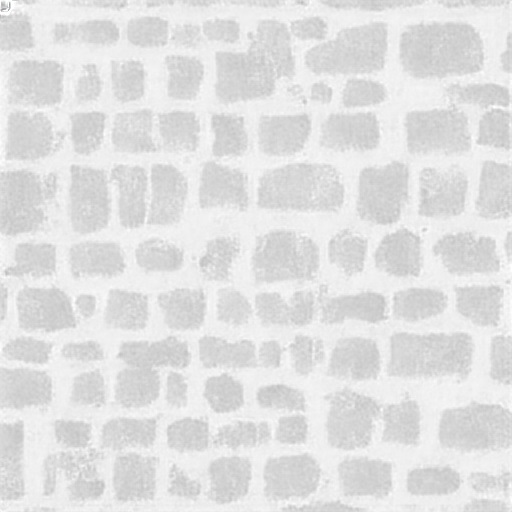} &
        \includegraphics[width=0.166\linewidth]{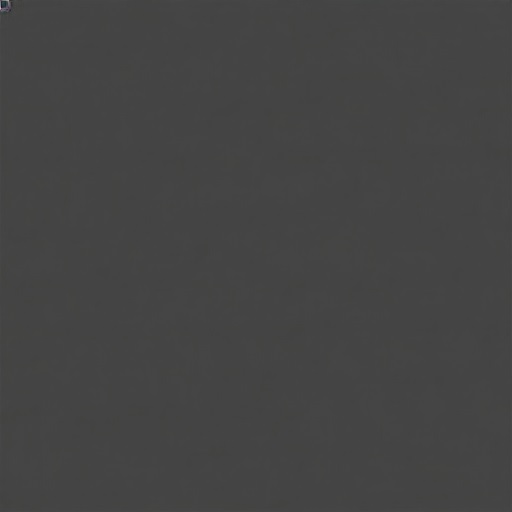} &
        \includegraphics[width=0.166\linewidth]{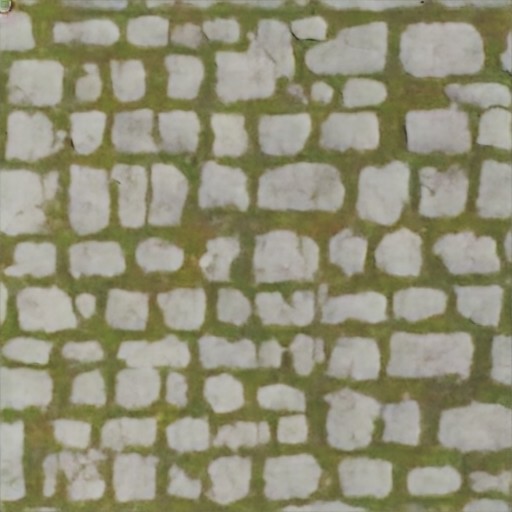} \\

        \vspace{-1mm}\hspace{-1mm}\includegraphics[width=0.166\linewidth]{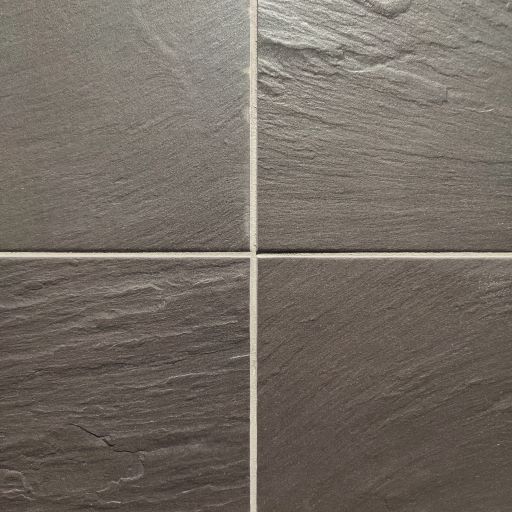} &
        \includegraphics[width=0.166\linewidth]{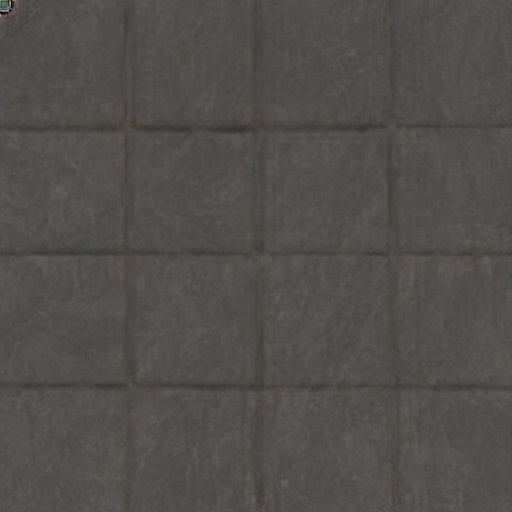} &
        \includegraphics[width=0.166\linewidth]{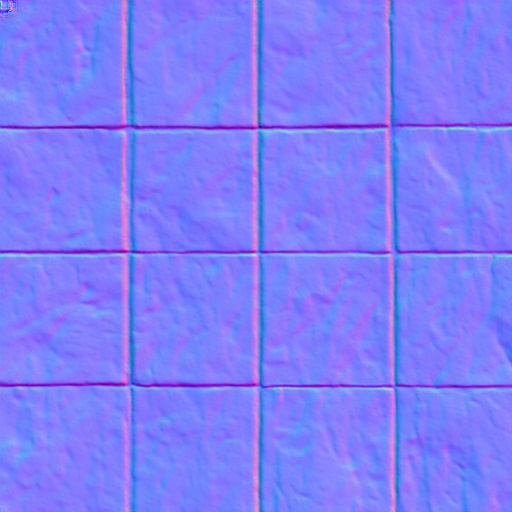} &
        \includegraphics[width=0.166\linewidth]{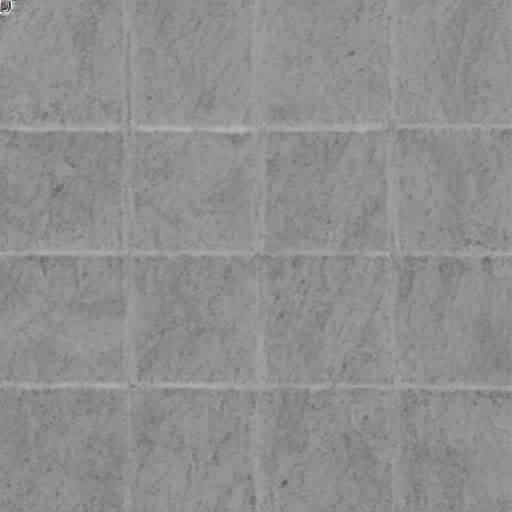} &
        \includegraphics[width=0.166\linewidth]{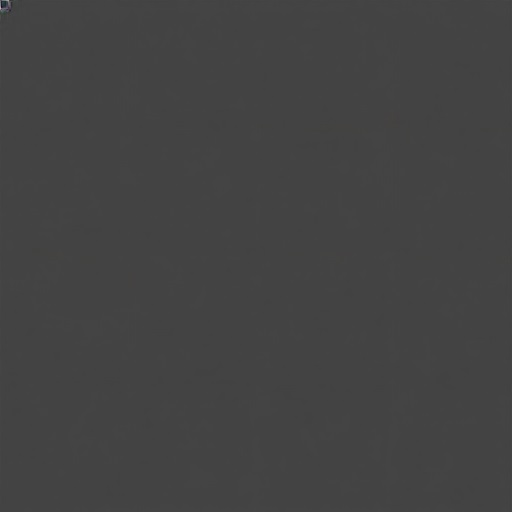} &
        \includegraphics[width=0.166\linewidth]{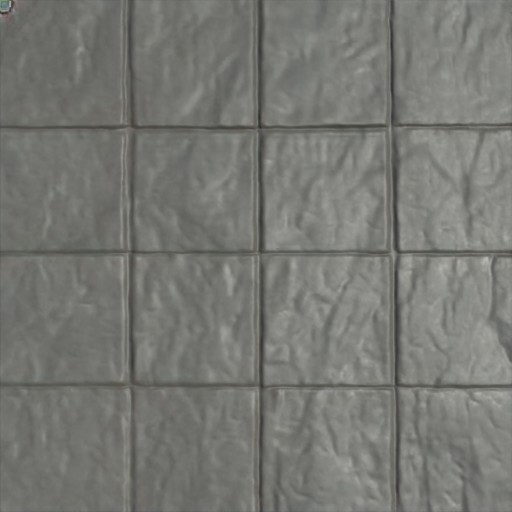} \\
    
        \vspace{-1mm}\hspace{-1mm}\includegraphics[width=0.166\linewidth,height=0.166\linewidth]{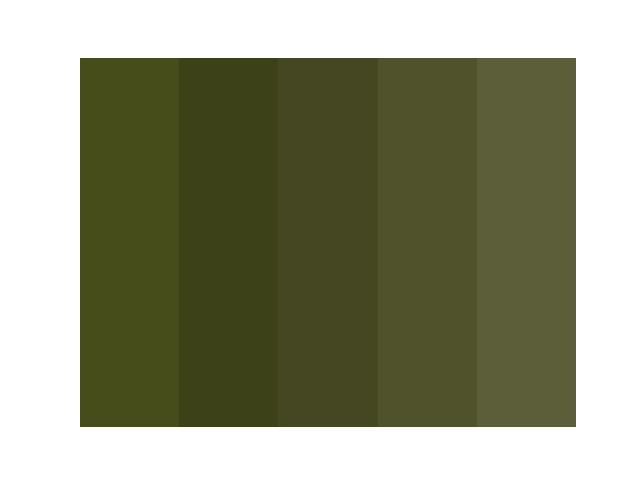} &
        \includegraphics[width=0.166\linewidth]{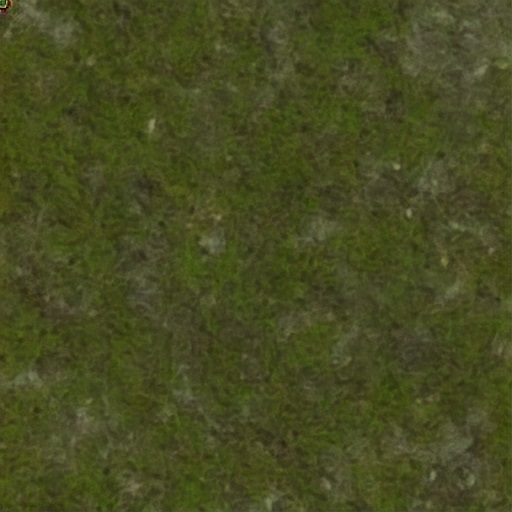} &
        \includegraphics[width=0.166\linewidth]{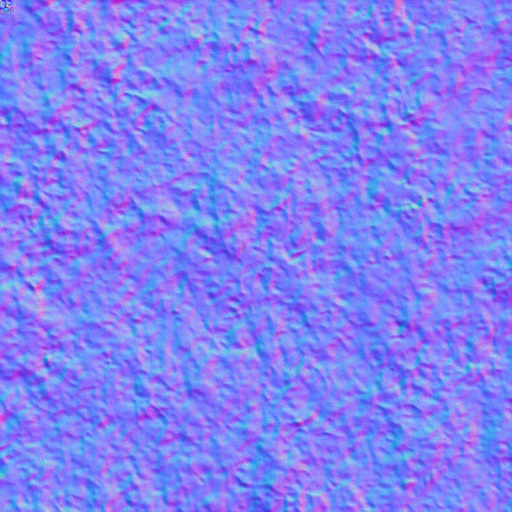} &
        \includegraphics[width=0.166\linewidth]{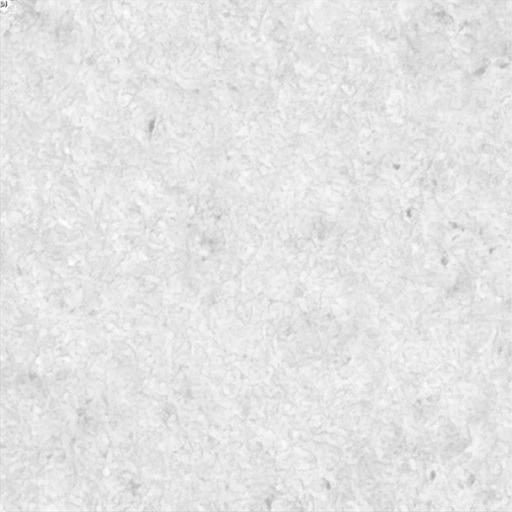} &
        \includegraphics[width=0.166\linewidth]{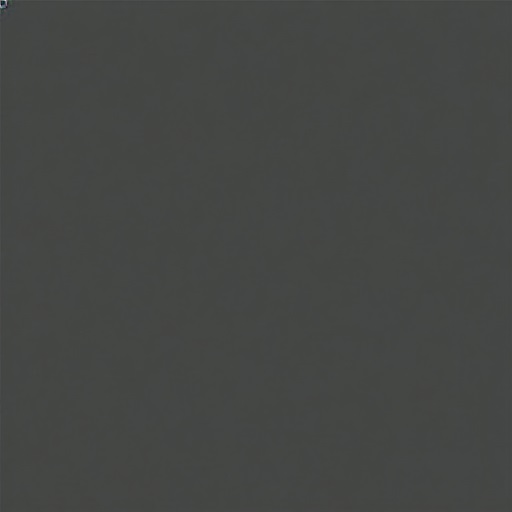} &
        \includegraphics[width=0.166\linewidth]{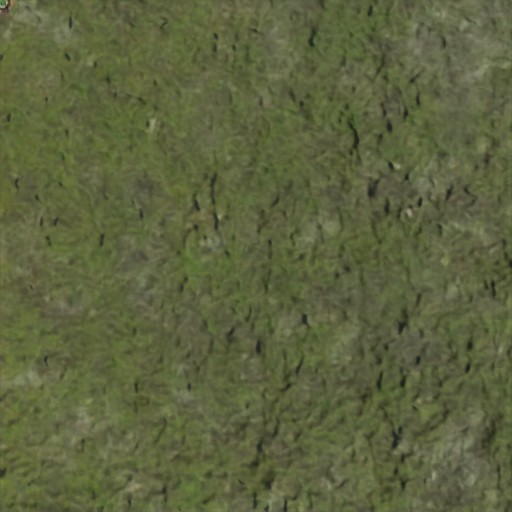} \\
    
        \vspace{-1mm}\hspace{-1mm}\includegraphics[width=0.166\linewidth,height=0.166\linewidth]{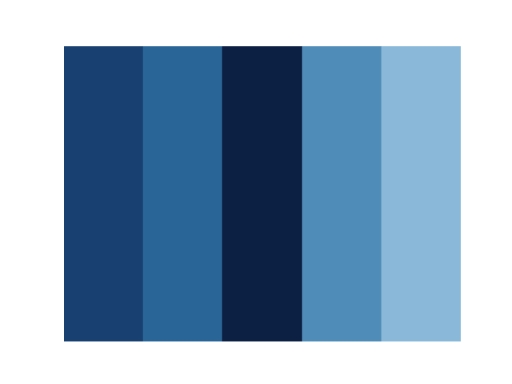} &
        \includegraphics[width=0.166\linewidth]{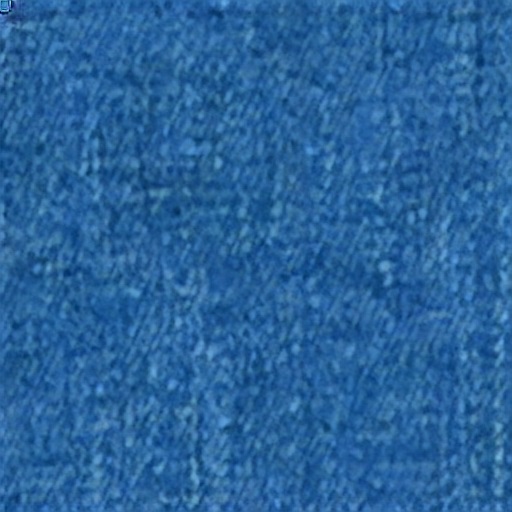} &
        \includegraphics[width=0.166\linewidth]{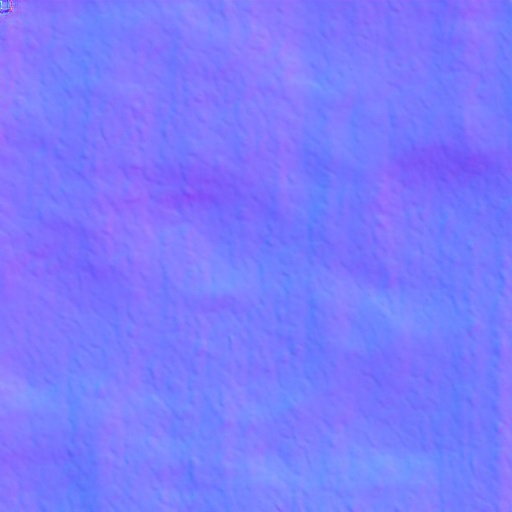} &
        \includegraphics[width=0.166\linewidth]{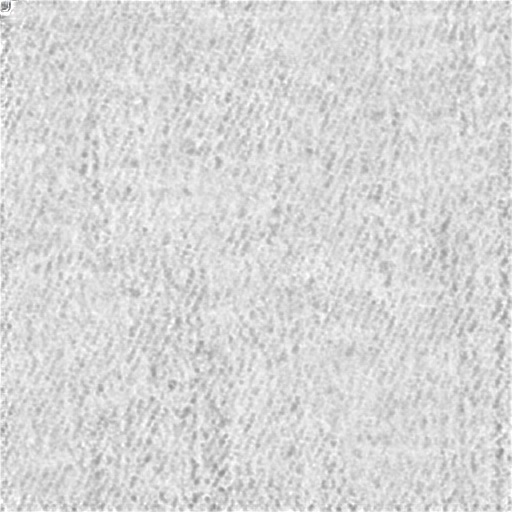} &
        \includegraphics[width=0.166\linewidth]{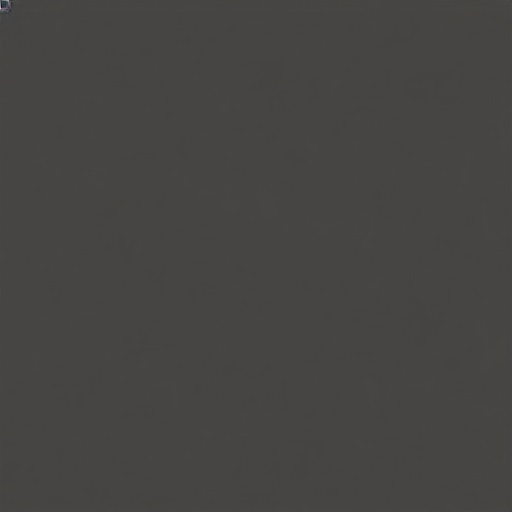} &
        \includegraphics[width=0.166\linewidth]{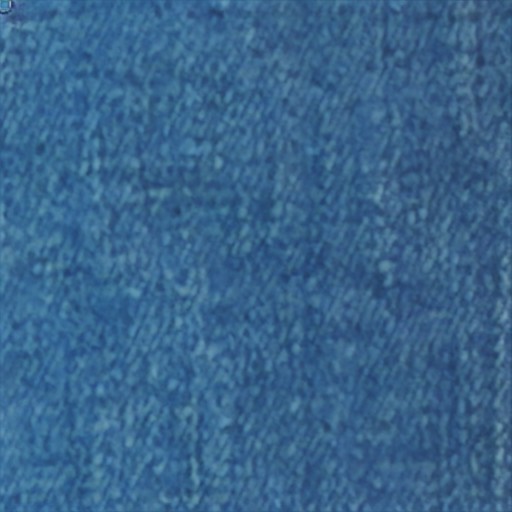} \\

    \end{tabular}
    
    \caption{\textbf{Globally conditioned material generation}. We evaluate \modelname when guided with single conditions. First two rows: text-conditioned map generation; mid two rows: image-prompted generation, yielding maps with features of the input image; last two rows: palette-conditioned generation.}
    \label{fig:gen_global}
\end{figure}

\noindent \textbf{Global conditioning}.
\modelname can be globally conditioned via text prompts, image prompts, and a color palette. Fig.~\ref{fig:gen_global} shows that our approach successfully captures the condition features in the generated materials. It is worth noting that the use of adjectives like ``shiny'' in the text prompt alters the visual appearance accordingly, for example by making the material less rough (Fig.~\ref{fig:gen_global}, second row).
Similarly, visual features from image prompts, like the highlight in the stone tiles, are correctly captured, as it is clearly visible in the resulting rendering.
Finally, colors from the color palette are accurately reproduced in the diffuse component of the generated material.

\begin{figure}
    \centering
    \setlength{\tabcolsep}{.5pt}
    \begin{tabular}{cccccc}
        \hspace{-1mm}\small{Condition} & \small{Diffuse} & \small{Normal} & \small{Roughness} & \small{Specular} & \small{Render} \\
        
        \vspace{-1mm}\hspace{-1mm}\includegraphics[width=0.166\linewidth]{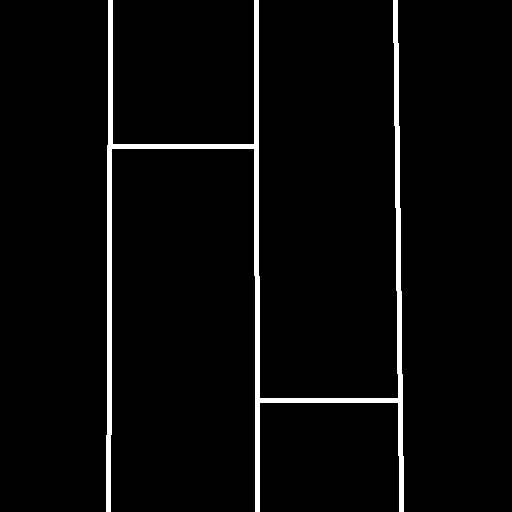} &
        \includegraphics[width=0.166\linewidth]{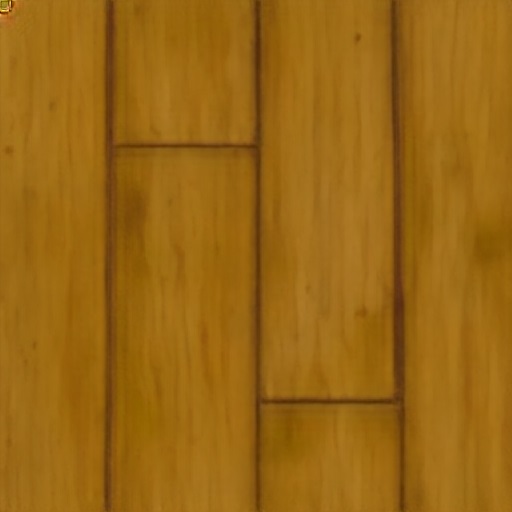} &
        \includegraphics[width=0.166\linewidth]{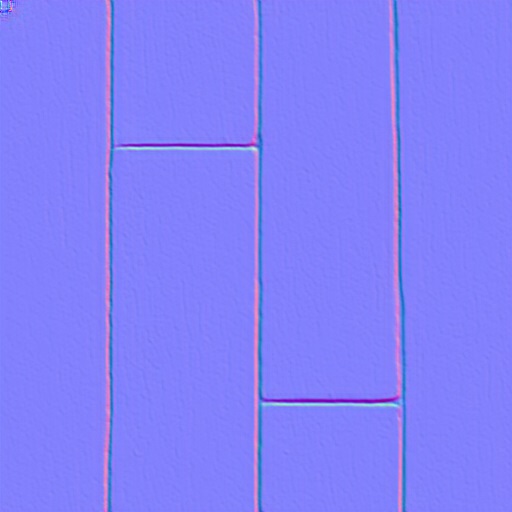} &
        \includegraphics[width=0.166\linewidth]{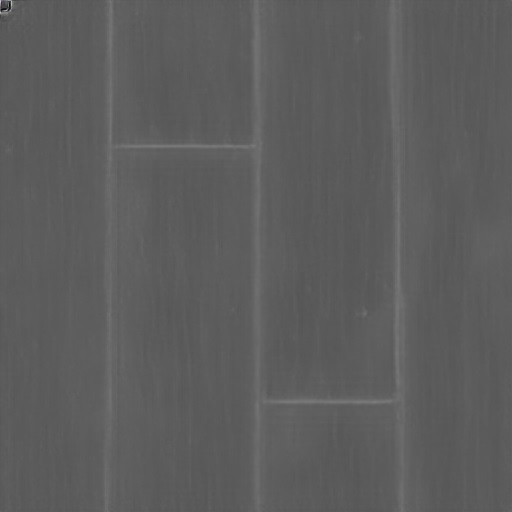} &
        \includegraphics[width=0.166\linewidth]{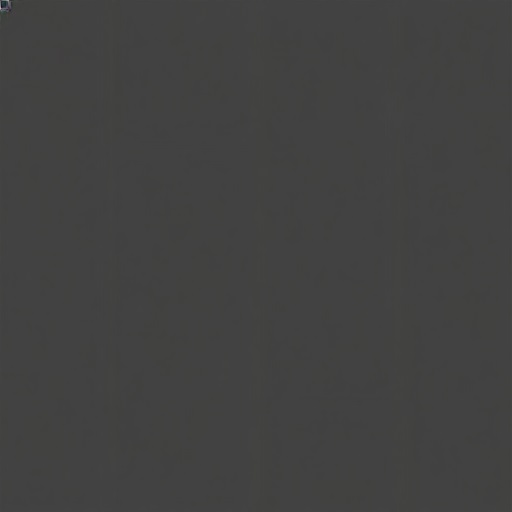} &
        \includegraphics[width=0.166\linewidth]{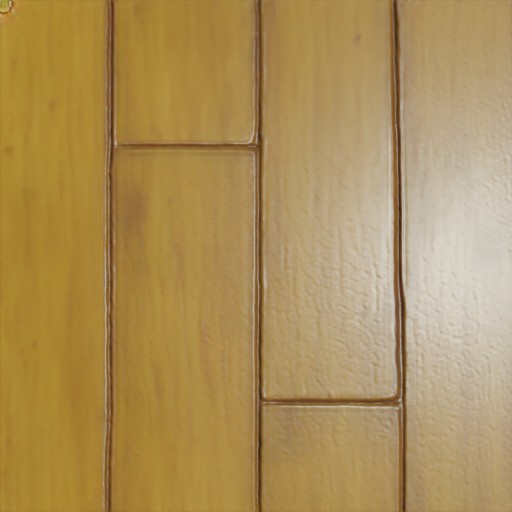} \\
    
        \vspace{-1mm}\hspace{-1mm}\includegraphics[width=0.166\linewidth]{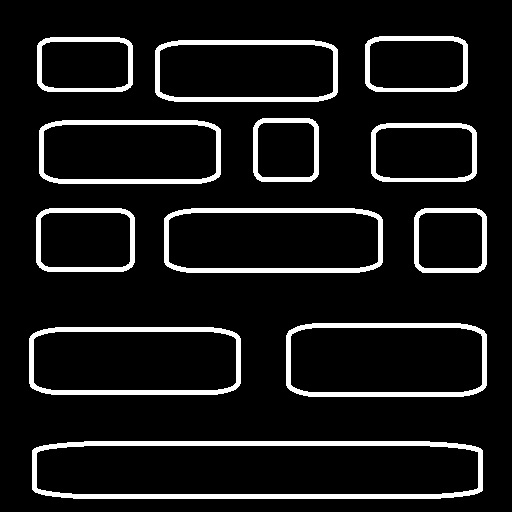} &
        \includegraphics[width=0.166\linewidth]{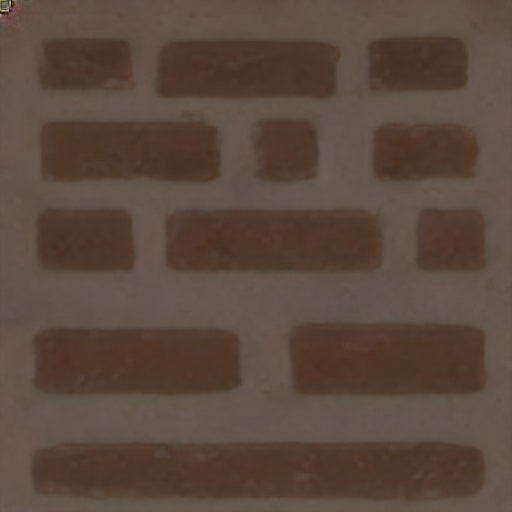} &
        \includegraphics[width=0.166\linewidth]{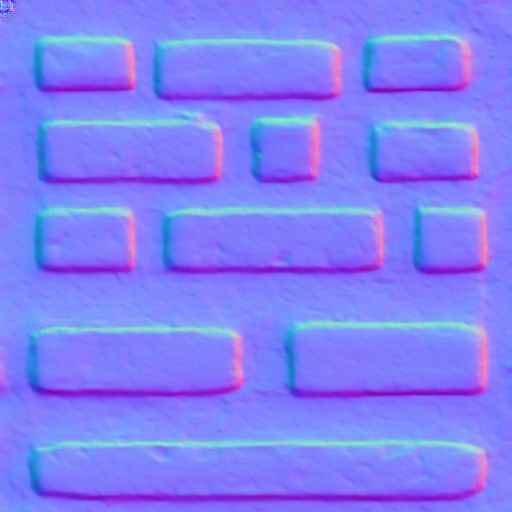} &
        \includegraphics[width=0.166\linewidth]{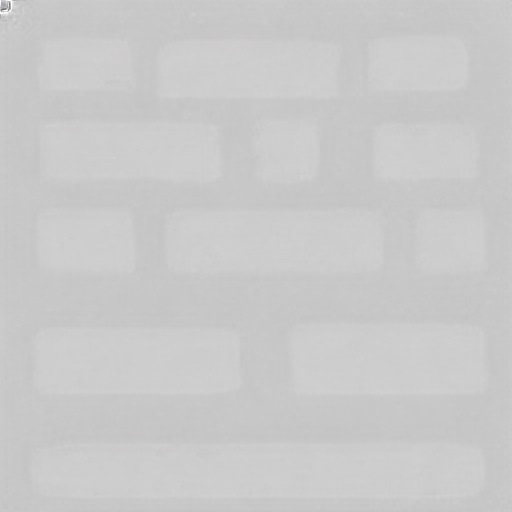} &
        \includegraphics[width=0.166\linewidth]{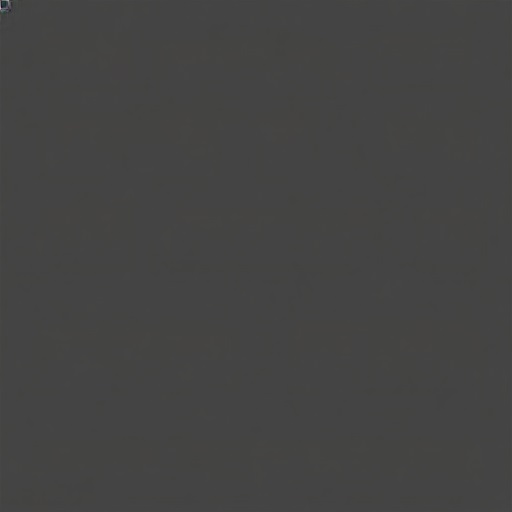} &
        \includegraphics[width=0.166\linewidth]{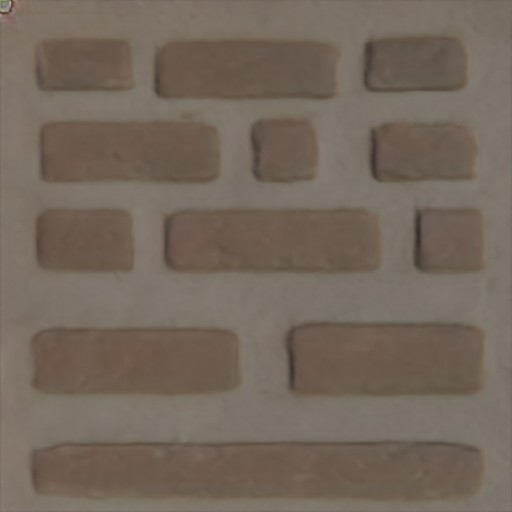} \\
    
        \vspace{-1mm}\hspace{-1mm}\includegraphics[width=0.166\linewidth]{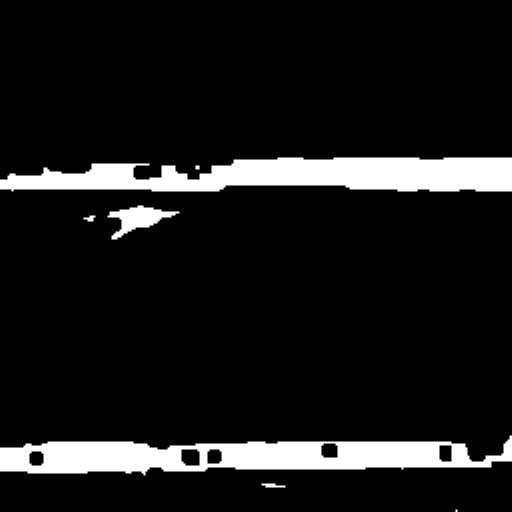} &
        \includegraphics[width=0.166\linewidth]{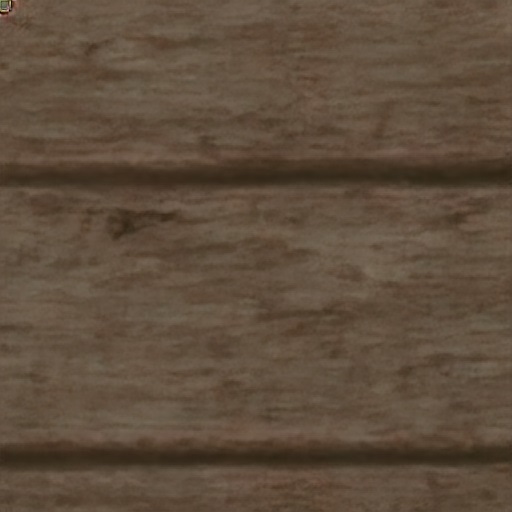} &
        \includegraphics[width=0.166\linewidth]{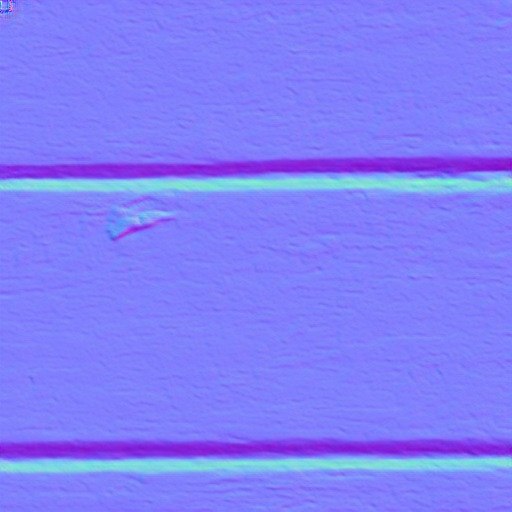} &
        \includegraphics[width=0.166\linewidth]{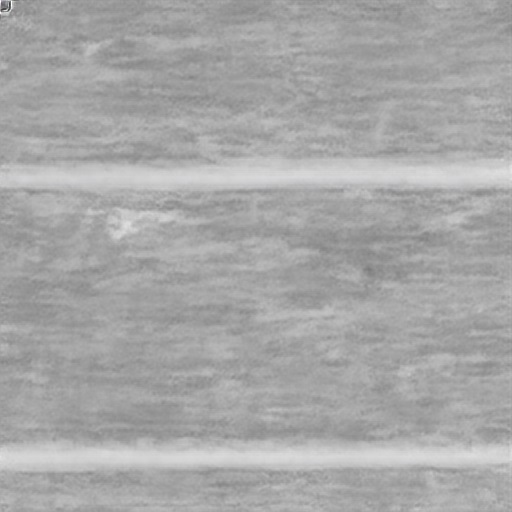} &
        \includegraphics[width=0.166\linewidth]{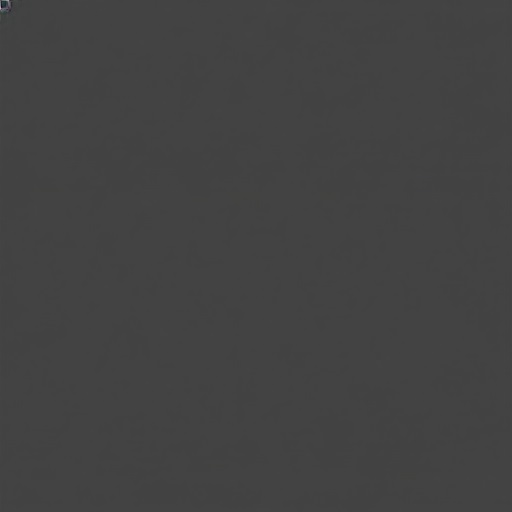} &
        \includegraphics[width=0.166\linewidth]{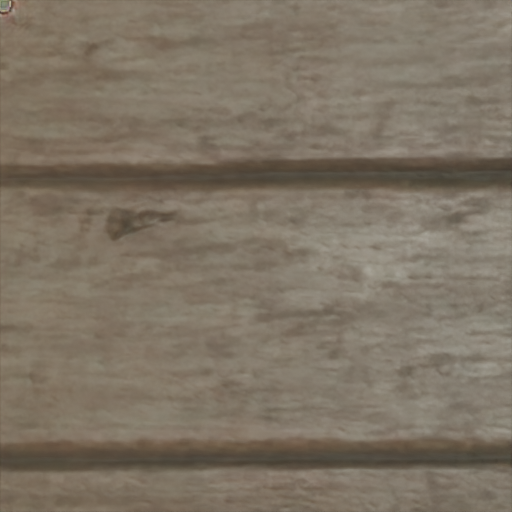} \\
    
    \end{tabular}
    
    \caption{\textbf{Locally conditioned material generation}. We provide sketches to condition \modelname and produce maps with well-defined edges. The first two rows present hand-drawn sketches, while the latter is obtained from a material picture. This shows the robustness of \modelname in handling both clean and noisy sketches.}
    \label{fig:gen_local}
\end{figure}

\noindent \textbf{Local conditioning}. 
\modelname can be locally conditioned through pattern sketches as demonstrated in Fig.~\ref{fig:gen_local}. 
The model can correctly process hand-drawn, clean, sketches as well as noisy sketches generated using a Canny edge detector. The model correctly transfers the structure defined in the sketch to the produced material by acting on its normal map, providing it with the desired geometry. 

\subsubsection{Multi-conditional generation}
\begin{figure}
    \centering
    \setlength{\tabcolsep}{.5pt}
    \begin{tabular}{ccccccc}       
        \hspace{-1mm}\small{Cond.} & \small{Sketch} & \small{Diffuse} & \small{Normal} & \small{Rough.} & \small{Specular} & \small{Render} \\
        
        \vspace{-1mm}\hspace{-1mm}\makecell[b]{\footnotesize{``rusty} \\ \footnotesize{metal''}\vspace{1.8mm}} &
        \includegraphics[width=0.142\linewidth, height=0.142\linewidth]{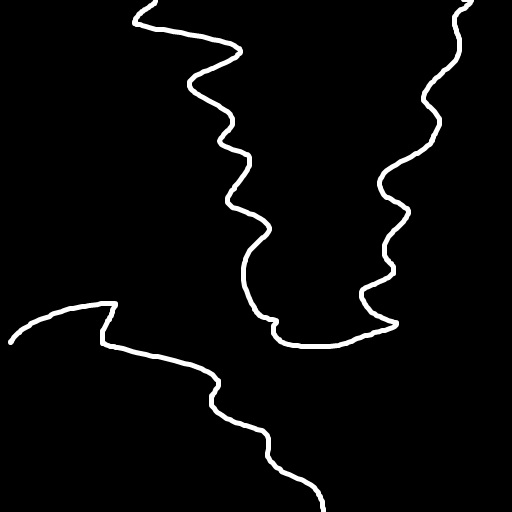} &
        \includegraphics[width=0.142\linewidth, height=0.142\linewidth]{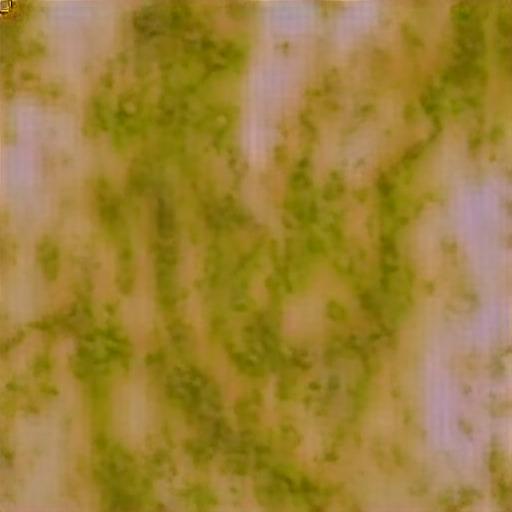} &
        \includegraphics[width=0.142\linewidth, height=0.142\linewidth]{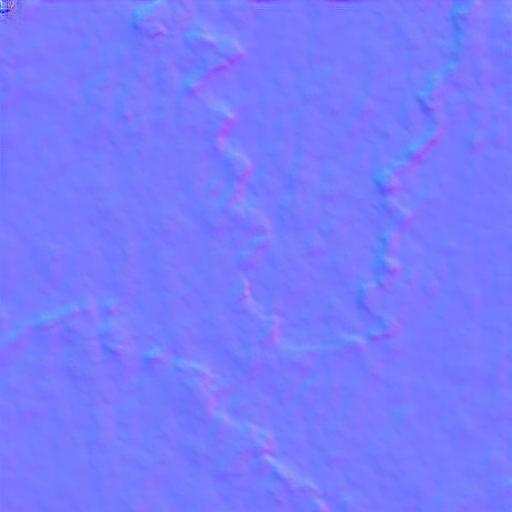} &
        \includegraphics[width=0.142\linewidth, height=0.142\linewidth]{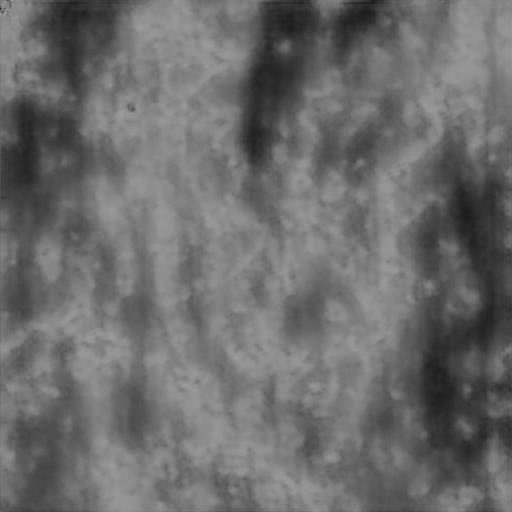} &
        \includegraphics[width=0.142\linewidth, height=0.142\linewidth]{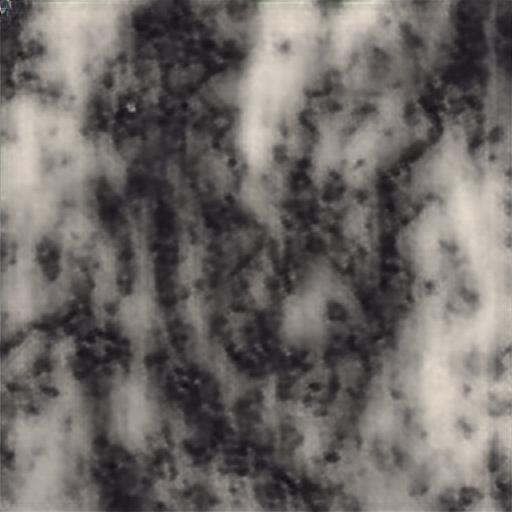} &
        \includegraphics[width=0.142\linewidth, height=0.142\linewidth]{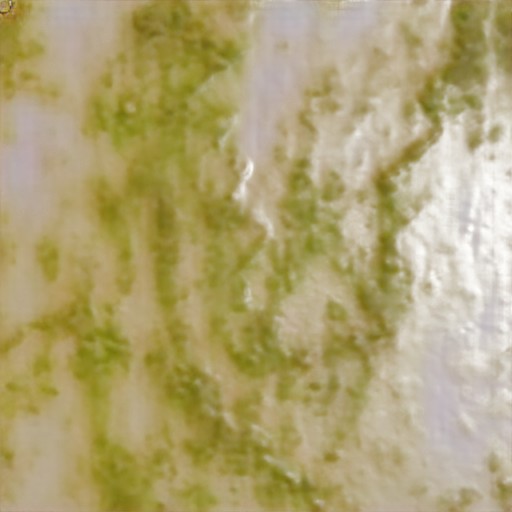} \\
    
        \vspace{-1mm}\hspace{-1mm}\includegraphics[width=0.142\linewidth, height=0.142\linewidth]{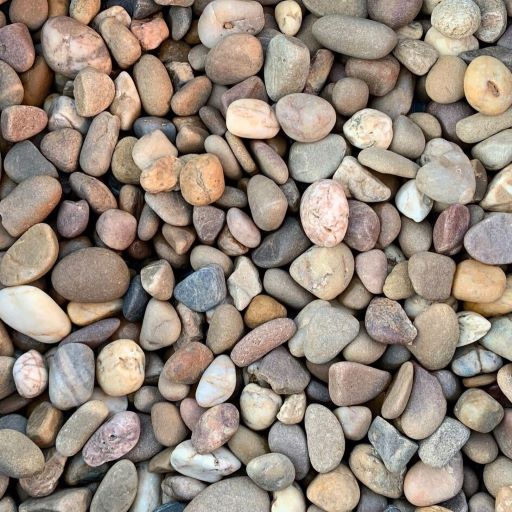} &
        \includegraphics[width=0.142\linewidth, height=0.142\linewidth]{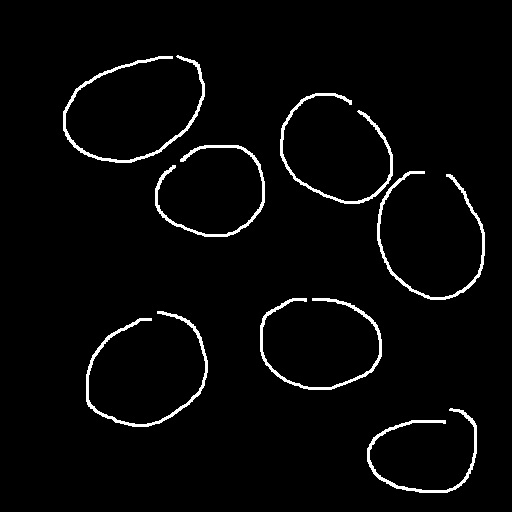} &
        \includegraphics[width=0.142\linewidth, height=0.142\linewidth]{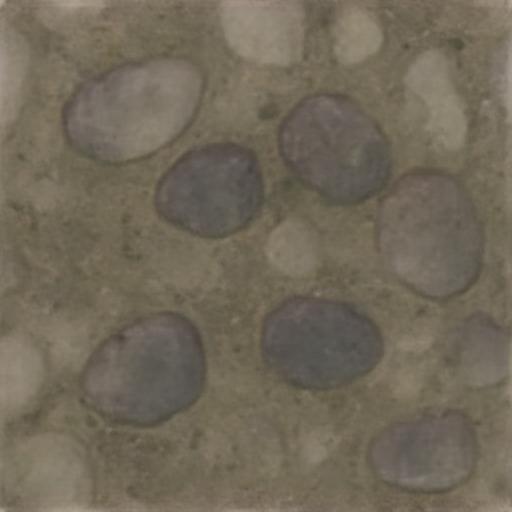} &
        \includegraphics[width=0.142\linewidth, height=0.142\linewidth]{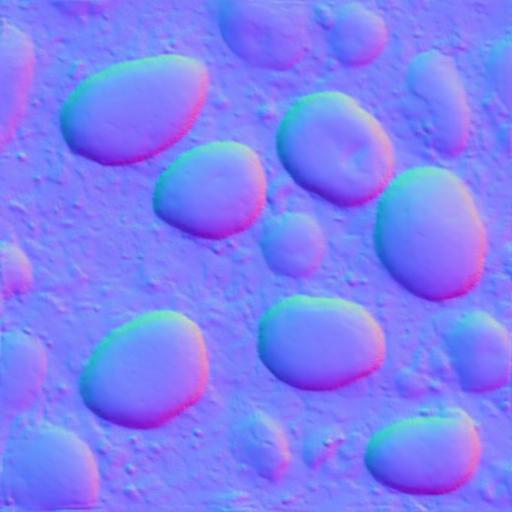} &
        \includegraphics[width=0.142\linewidth, height=0.142\linewidth]{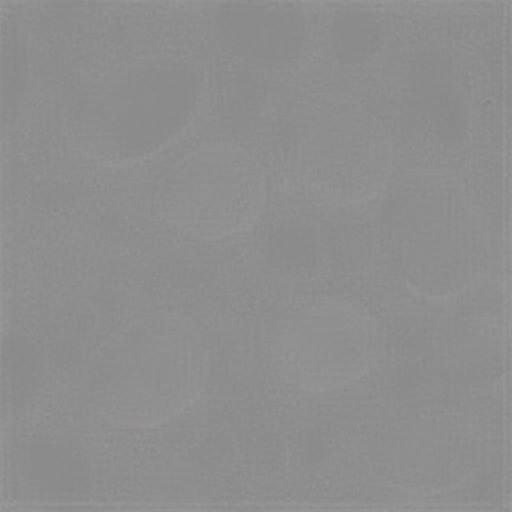} &
        \includegraphics[width=0.142\linewidth, height=0.142\linewidth]{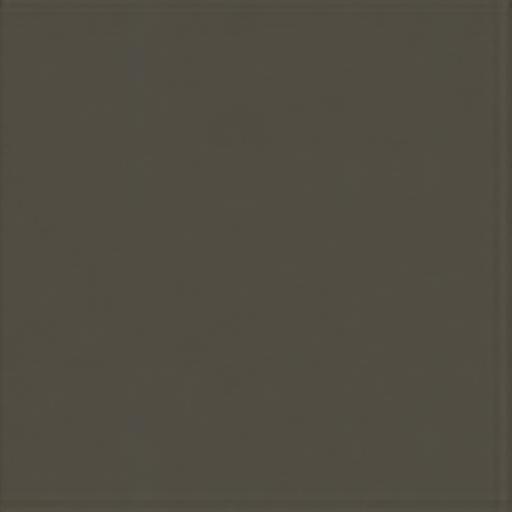} &
        \includegraphics[width=0.142\linewidth, height=0.142\linewidth]{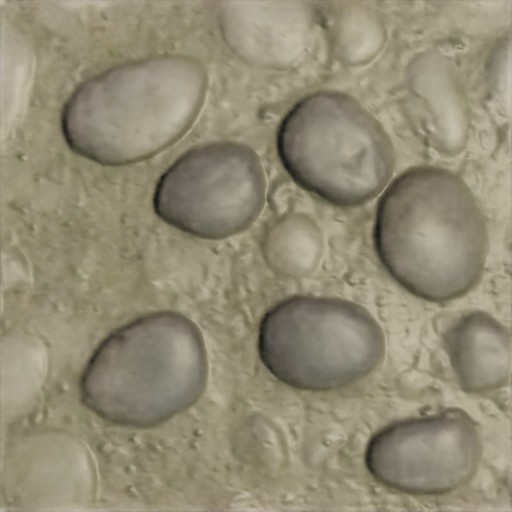} \\
    
        \vspace{-1mm}\hspace{-1mm}\includegraphics[width=0.142\linewidth, height=0.142\linewidth]{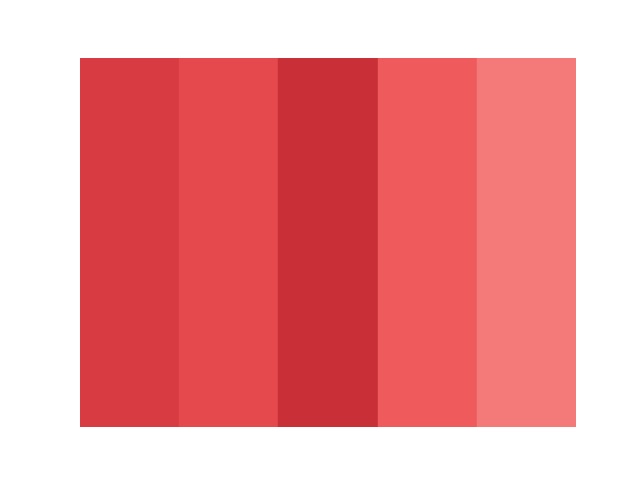} &
        \includegraphics[width=0.142\linewidth, height=0.142\linewidth]{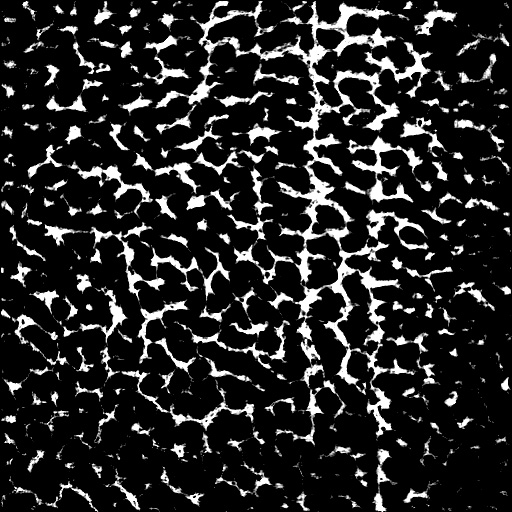} &
        \includegraphics[width=0.142\linewidth, height=0.142\linewidth]{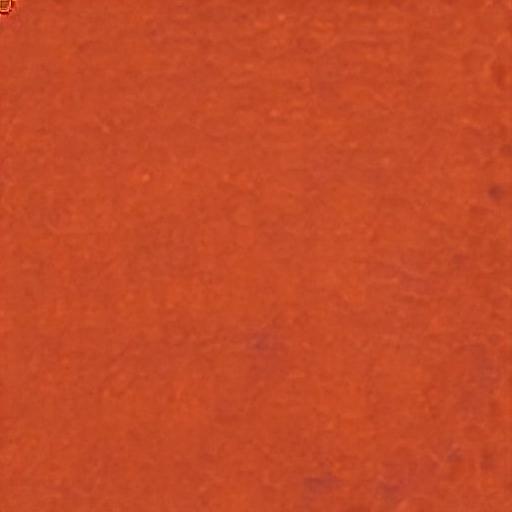} &
        \includegraphics[width=0.142\linewidth, height=0.142\linewidth]{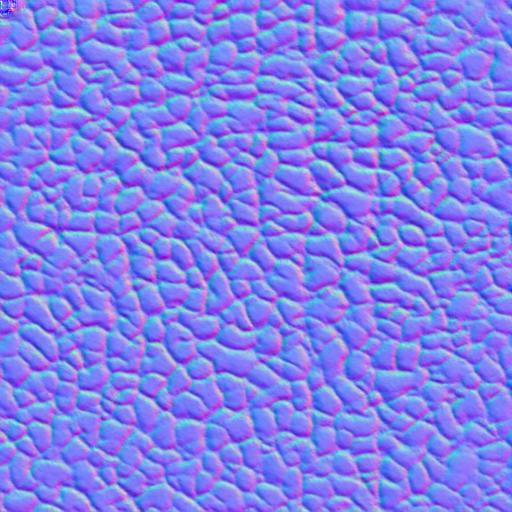} &
        \includegraphics[width=0.142\linewidth, height=0.142\linewidth]{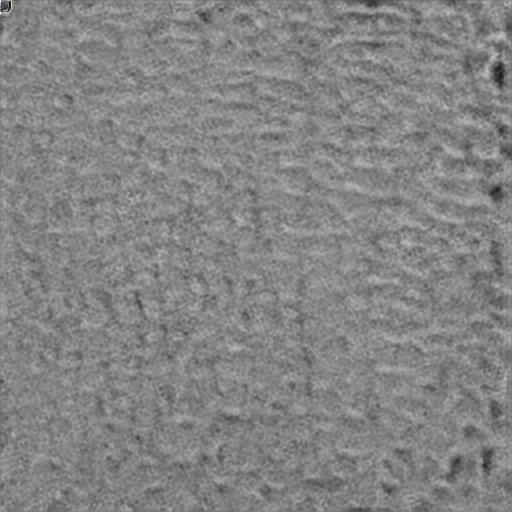} &
        \includegraphics[width=0.142\linewidth, height=0.142\linewidth]{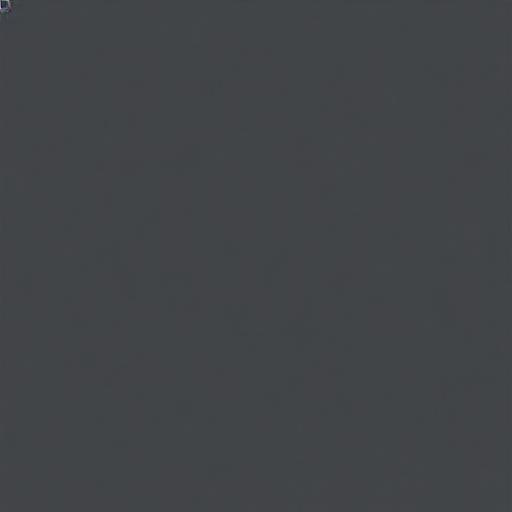} &
        \includegraphics[width=0.142\linewidth, height=0.142\linewidth]{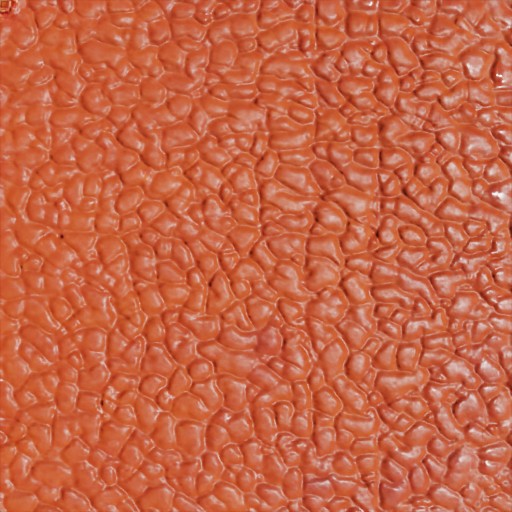} \\

    \end{tabular}
    
    \caption{\textbf{Multimodal conditioned material generation}. First row: text prompt + sketch. Second row: image prompt + sketch. Third row: color palette + sketch.}
    \label{fig:gen_multi}
\end{figure}

The main strength of \modelname resides in the possibility of combining multiple conditions for a finer generation control. 
In particular, combining a local and a global condition gives control over both the geometry and the visual features of the material.
Fig.~\ref{fig:gen_multi} shows the materials generated when combining the sketch with each of the global conditions. We can see that the model is able to accurately follow the spatial structure given by the sketch while showing the semantics of the global conditions. 

\subsection{Quantitative Evaluation} 
Assessing material generation quality is a challenging task, due to their inherently different data distribution compared to natural images. Established metrics such as the Fréchet Inception Distance (FID)~\cite{heusel2017gans} or the Inception Score~\cite{salimans2016improved} rely on the InceptionV3~\cite{szegedy2014going} architecture pre-trained on ImageNet~\cite{deng2009imagenet}, which includes natural images. 

To evaluate our results we employ the CLIP-IQA, recently introduced by \citet{wang2023exploring}.
This approach uses contrastive prompts to determine the appearance of an image. In particular, we evaluated the quality of the generated materials using the ``high-quality/low-quality'' contrastive pair. We believe that this metric is more suitable for the task at hand due to the much wider range of data used to train the CLIP model~\cite{clip}.

To this aim, we render 2,000 unconditionally generated materials at 512$\times$512 resolution, and measure both the CLIP-IQA and FID on these renders. For reference, we also compute the CLIP-IQA on the ground truth renders from our dataset, as well on samples generated by vanilla LDM~\cite{rombach2022high} and TileGen~\cite{zhou2022tilegen}\footnote{As the model has not been publicly released, a batch of samples generated by TileGen was kindly provided by the authors.}.
Tab.~\ref{tab:iqa} reports the results of this analysis. On the CLIP-IQA metric, \modelname improves significantly over the baseline model and produces scores close to the target upper bound, i.e., ground truth samples from the employed dataset. Notably, \modelname is almost on par with TileGen~\cite{zhou2022tilegen}, although the latter is trained on a more restricted set of classes. FID scores confirm a higher similarity between \modelname's samples and ground-truth images, compared to the baseline. It is interesting to note that, in this metric, \modelname significantly outperforms TileGen too, which is likely due to TileGen's limited sample diversity, captured by the FID score.

\begin{table}
    \centering
    \begin{tabular}{lcc}
    \toprule
        Model  & CLIP-IQA ($\uparrow$) & FID ($\downarrow$) \\
        \midrule
        Ground Truth & 0.471 $\pm$ 0.191 & - \\
        \midrule
        TileGen~\cite{zhou2022tilegen} & 0.433 $\pm$ 0.161 & 184.81 \\
        \midrule
        LDM (baseline) & 0.269 $\pm$ 0.118 & 231.64\\
        \textbf{\modelname} & \textbf{0.431} $\pm$ 0.151 & 158.53\\
       \bottomrule
    \end{tabular}
    \caption{\textbf{Performance of \modelname in terms of CLIP-IQA}. The CLIP-IQA values for the datasets used during the training serve as our upper bound. We compare the generation quality to TileGen~\cite{zhou2022tilegen} and a baseline Stable Diffusion model trained to generate materials unconditionally. The CLIP-IQA metric is computed using the ``high-quality/low-quality'' contrastive pair.}
    \label{tab:iqa}
\end{table}

\subsection{Qualitative Comparison and User Study}
\begin{figure}
    \centering
    \setlength{\tabcolsep}{.5pt}
    \hspace{2mm}\begin{tabular} {ccccccc}
     & & \small{Diffuse} & \small{Normal} & \small{Roughness} & \small{Specular} & \small{Render} \\
     
    \vspace{-1mm}\hspace{-6mm}\begin{sideways}\hspace{1mm} \small{TileGen} \end{sideways} &
    \multirow{2}{*}{\begin{sideways} \small{Leather} \end{sideways}} \hspace{2mm} &
    \includegraphics[width=0.166\linewidth, height=0.166\linewidth]{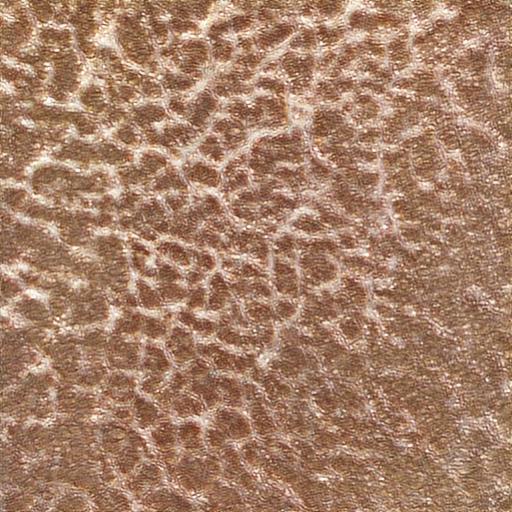} &
    \includegraphics[width=0.166\linewidth, height=0.166\linewidth]{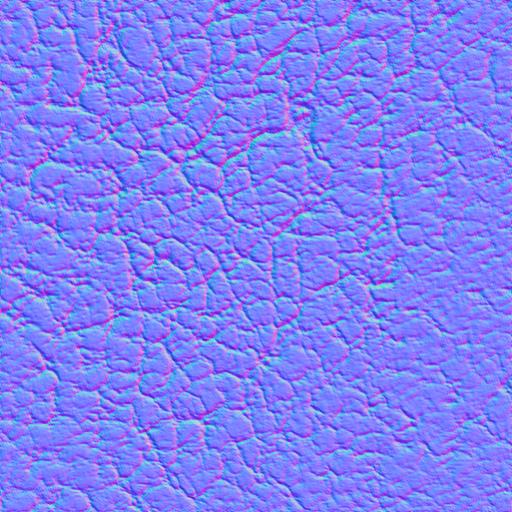} &
    \includegraphics[width=0.166\linewidth, height=0.166\linewidth]{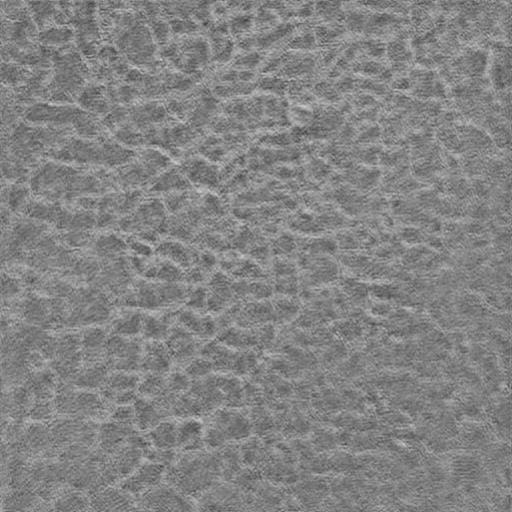} &
    \includegraphics[width=0.166\linewidth, height=0.166\linewidth]{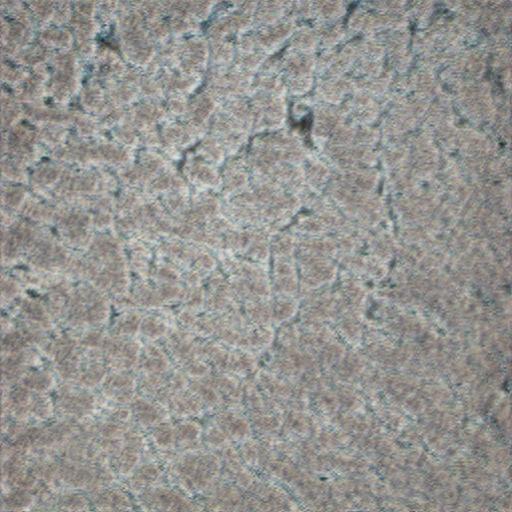} &
    \includegraphics[width=0.166\linewidth, height=0.166\linewidth]{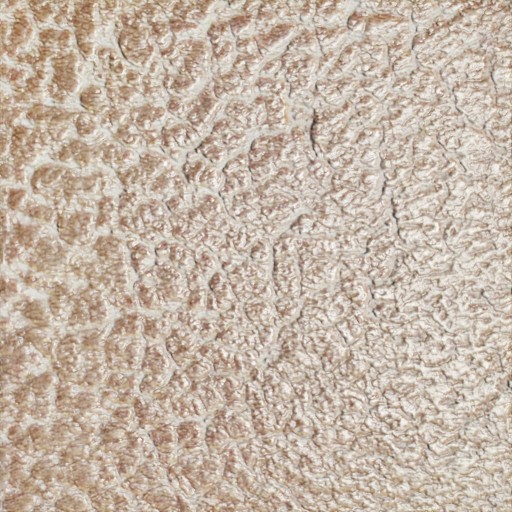} \\
    
    \vspace{-1mm}\hspace{-6mm}\begin{sideways}\hspace{1mm} \small{MatFuse} \end{sideways} & & 
    \includegraphics[width=0.166\linewidth, height=0.166\linewidth]{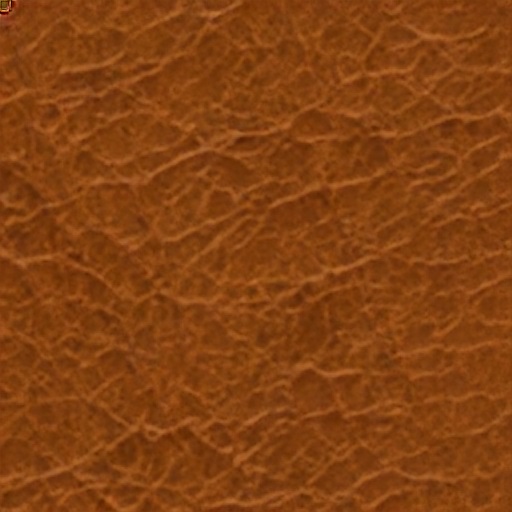} &
    \includegraphics[width=0.166\linewidth, height=0.166\linewidth]{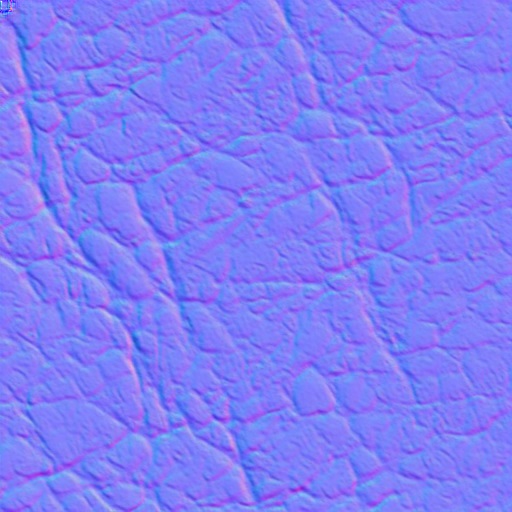} &
    \includegraphics[width=0.166\linewidth, height=0.166\linewidth]{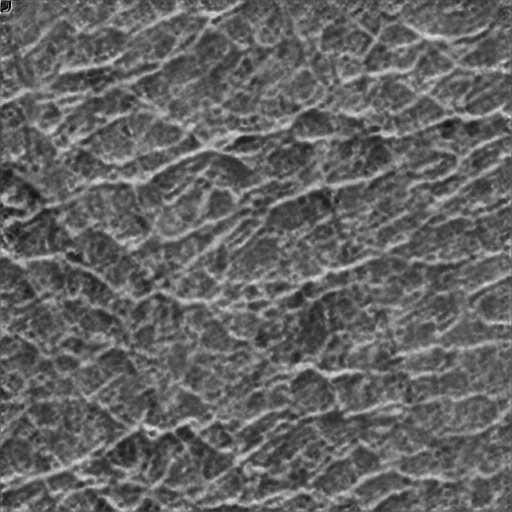} &
    \includegraphics[width=0.166\linewidth, height=0.166\linewidth]{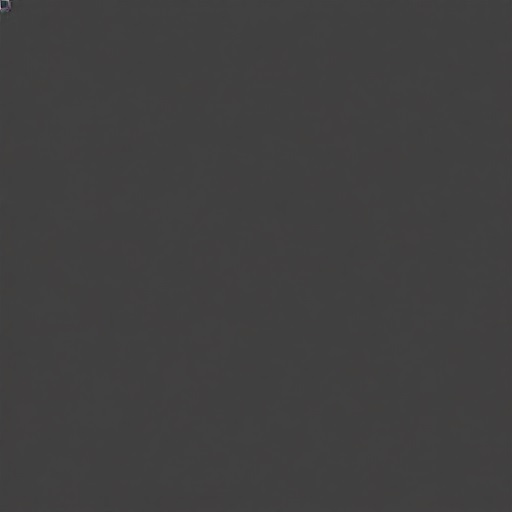} &
    \includegraphics[width=0.166\linewidth, height=0.166\linewidth]{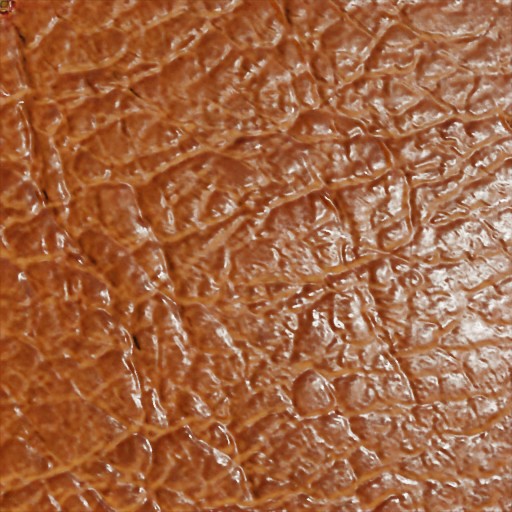} \\

    \vspace{-1mm}\hspace{-6mm}\begin{sideways}\hspace{1mm} \small{TileGen} \end{sideways} &
    \multirow{2}{*}{\begin{sideways} \small{Wood} \end{sideways}} \hspace{2mm} &
    \includegraphics[width=0.166\linewidth, height=0.166\linewidth]{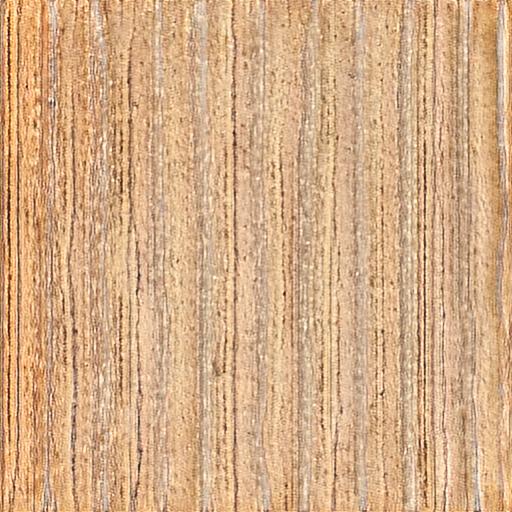} &
    \includegraphics[width=0.166\linewidth, height=0.166\linewidth]{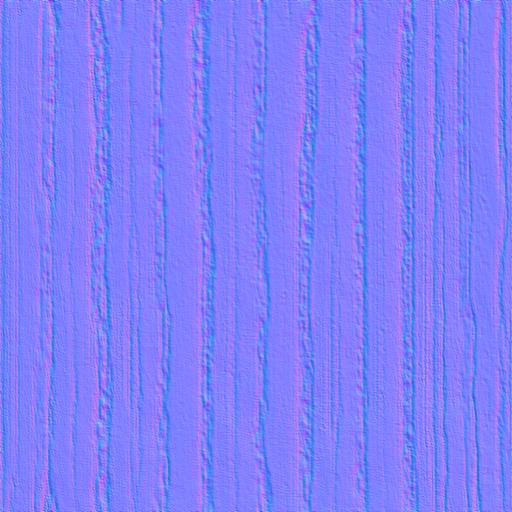} &
    \includegraphics[width=0.166\linewidth, height=0.166\linewidth]{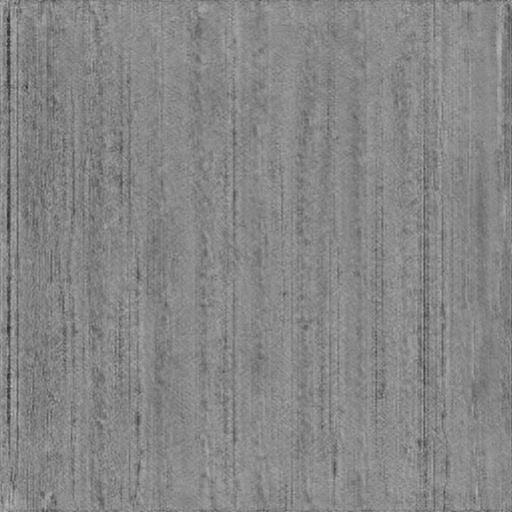} &
    \includegraphics[width=0.166\linewidth, height=0.166\linewidth]{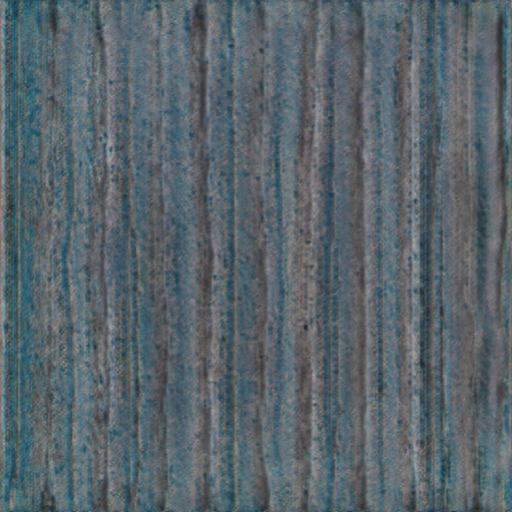} &
    \includegraphics[width=0.166\linewidth, height=0.166\linewidth]{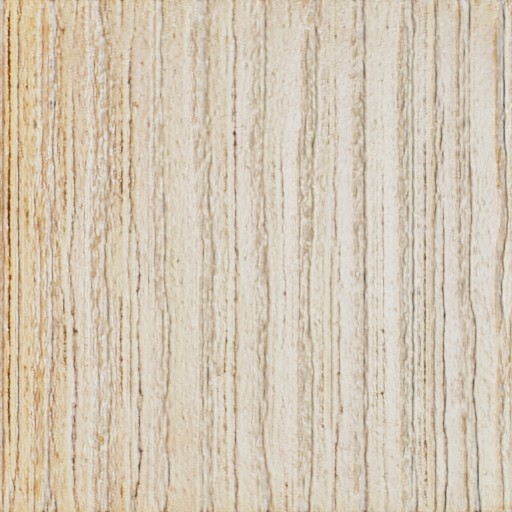} \\
    
    \vspace{-1mm}\hspace{-6mm}\begin{sideways}\hspace{1mm} \small{MatFuse} \end{sideways} & & 
    \includegraphics[width=0.166\linewidth, height=0.166\linewidth]{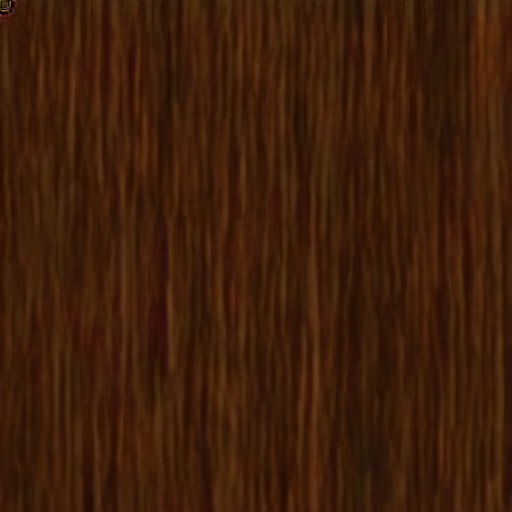} &
    \includegraphics[width=0.166\linewidth, height=0.166\linewidth]{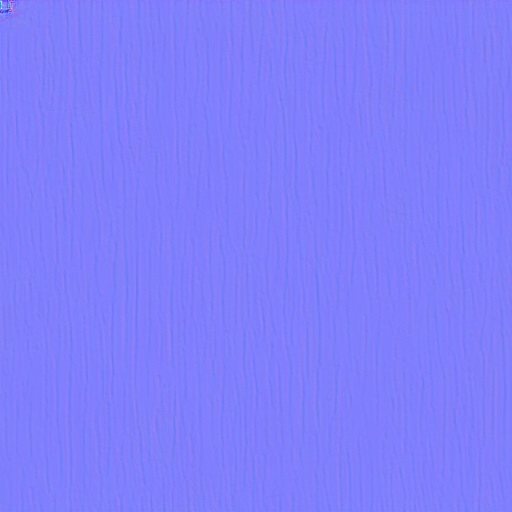} &
    \includegraphics[width=0.166\linewidth, height=0.166\linewidth]{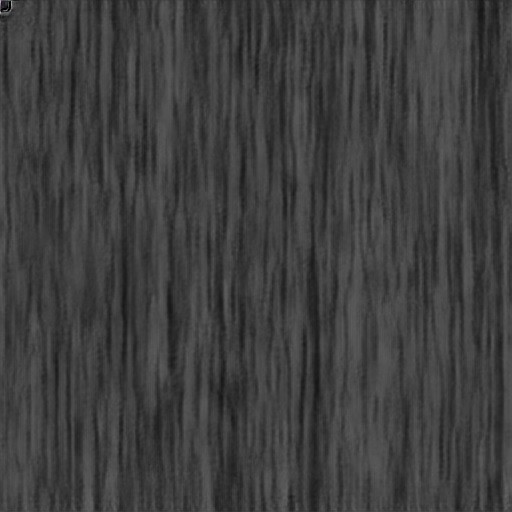} &
    \includegraphics[width=0.166\linewidth, height=0.166\linewidth]{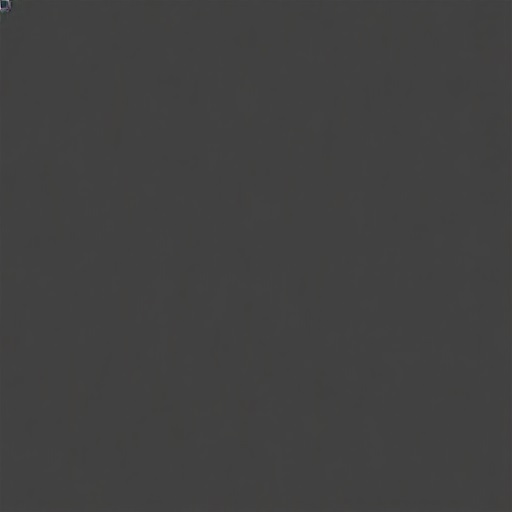} &
    \includegraphics[width=0.166\linewidth, height=0.166\linewidth]{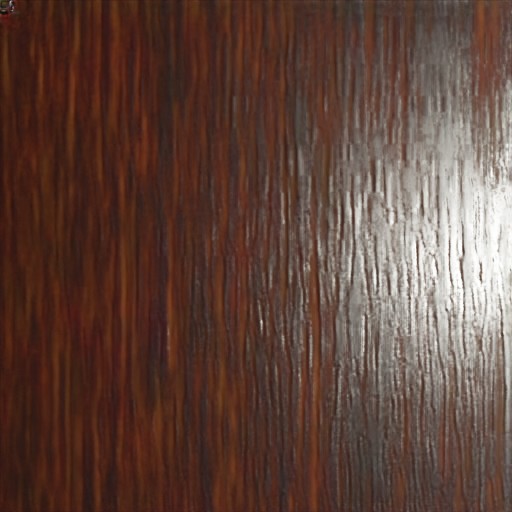} \\

    \vspace{-1mm}\hspace{-6mm}\begin{sideways}\hspace{1mm} \small{TileGen} \end{sideways} &
    \multirow{2}{*}{\begin{sideways} \small{Marble} \end{sideways}} \hspace{2mm} &
    \includegraphics[width=0.166\linewidth, height=0.166\linewidth]{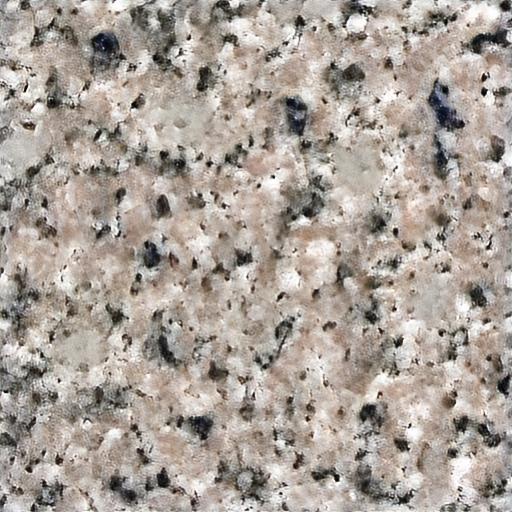} &
    \includegraphics[width=0.166\linewidth, height=0.166\linewidth]{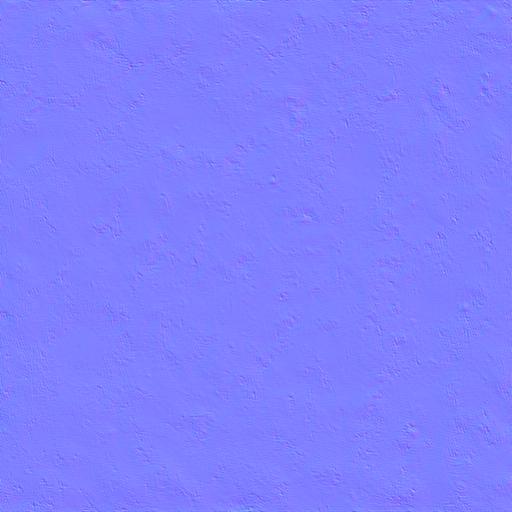} &
    \includegraphics[width=0.166\linewidth, height=0.166\linewidth]{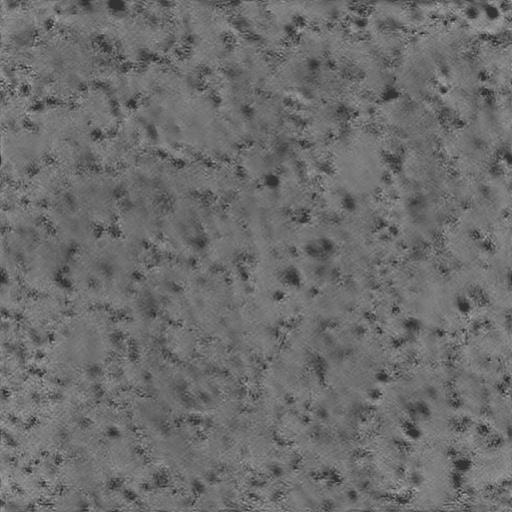} &
    \includegraphics[width=0.166\linewidth, height=0.166\linewidth]{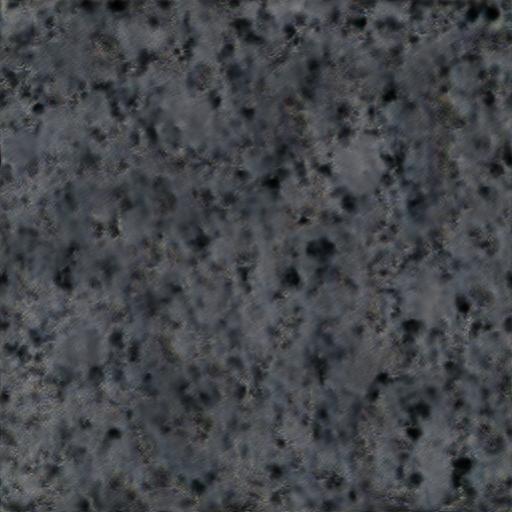} &
    \includegraphics[width=0.166\linewidth, height=0.166\linewidth]{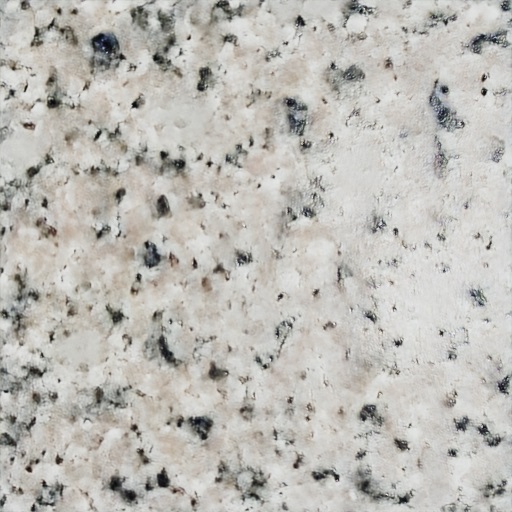} \\
    
    \vspace{-1mm}\hspace{-6mm}\begin{sideways}\hspace{1mm} \small{MatFuse} \end{sideways} & & 
    \includegraphics[width=0.166\linewidth, height=0.166\linewidth]{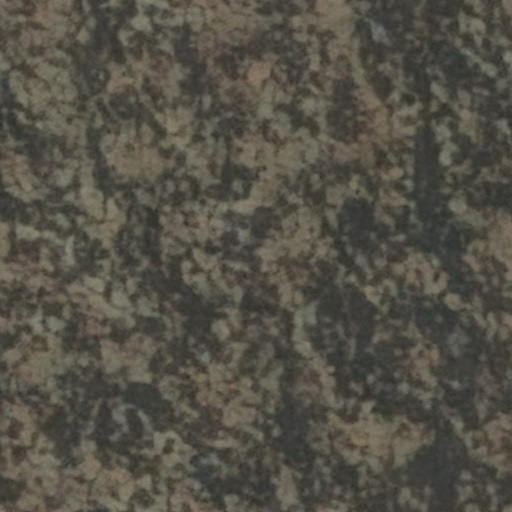} &
    \includegraphics[width=0.166\linewidth, height=0.166\linewidth]{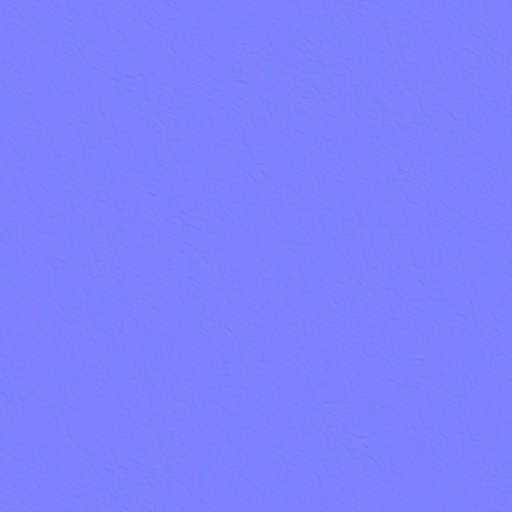} &
    \includegraphics[width=0.166\linewidth, height=0.166\linewidth]{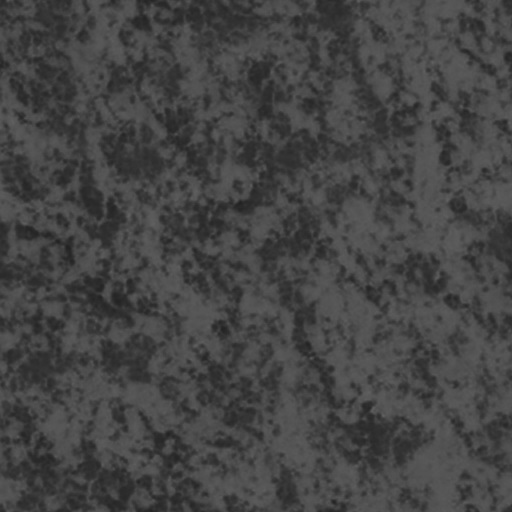} &
    \includegraphics[width=0.166\linewidth, height=0.166\linewidth]{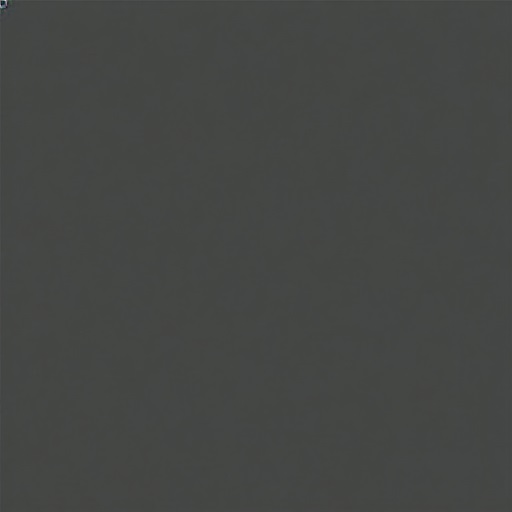} &
    \includegraphics[width=0.166\linewidth, height=0.166\linewidth]{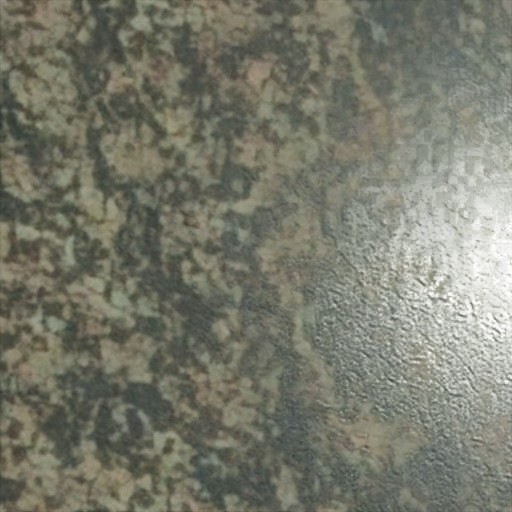} \\
    
    \end{tabular}
    \caption{\textbf{Comparison to TileGen~\cite{zhou2022tilegen}.} We compare MatFuse to TileGen models trained on three categories (Leather, Wood, Marble), conditioning our model with the category name. \suppmat{Additional samples are provided in the supplemental materials.}}
    \label{fig:comparison_tilegen}
\end{figure}

We further evaluated \modelname performances by conducting a pairwise comparison study between MatFuse and TileGen~\cite{zhou2022tilegen}. The study involved 100 MS/PhD students in computer science who were asked to choose their preferred materials based on both realism and rendering quality. We presented them with 25 randomly selected material pairs, drawn from a larger pool of 100 samples from the ``leather'', ``wood'', ``marble'', ``stone'' and ``ceramic'' classes. These classes were already available in TileGen and were used for conditional generation in MatFuse. 
Visual samples from the user study are presented in Figure~\ref{fig:comparison_tilegen}. Our method employs global conditioning by embedding class names as a condition for generation. 

In our test, MatFuse received a higher number of votes (MatFuse=1078, TileGen=949, No Pref.=473), and a chi-square test establishes statistical significance in the user's preference for MatFuse over TileGen ($\chi^2=$16.41, $p<0.05$).
It is important to note that both our method and TileGen generate results based solely on the specified class, leading to different appearances. Although \modelname is not specifically trained for a semantic class as opposed to TileGen, it is capable of producing high-quality materials with fine detail and a realistic appearance.

\subsection{Material Editing Results}
\label{sec:results_edit}
\begin{figure*}
    \centering
    \setlength{\tabcolsep}{.5pt}
    \begin{tabular}{cccccccccc}
        \hspace{-1mm} & \multicolumn{4}{c}{Input maps} & \multicolumn{4}{c}{Edited maps} \\
        \hspace{-1mm} \small{Condition} & \small{Diffuse} & \small{Normal} & \small{Roughness} & \small{Specular} & \small{Diffuse} & \small{Normal} & \small{Roughness} & \small{Specular} & \small{Render} \\
        
        \vspace{-1mm}\hspace{-1mm}\includegraphics[width=0.097\linewidth, height=0.097\linewidth]{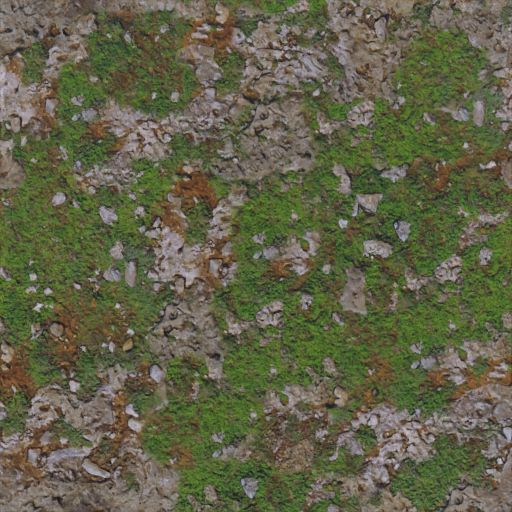} &
        \includegraphics[width=0.097\linewidth, height=0.097\linewidth]{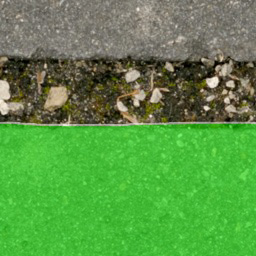} &
        \includegraphics[width=0.097\linewidth, height=0.097\linewidth]{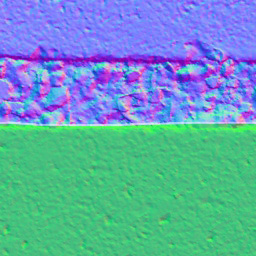} &
        \includegraphics[width=0.097\linewidth, height=0.097\linewidth]{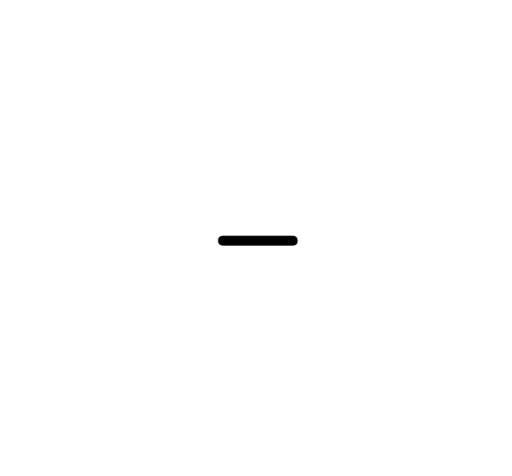} &
        \includegraphics[width=0.097\linewidth, height=0.097\linewidth]{figures/edit/placeholder.jpg} &
        \hspace{.5mm} \includegraphics[width=0.097\linewidth, height=0.097\linewidth]{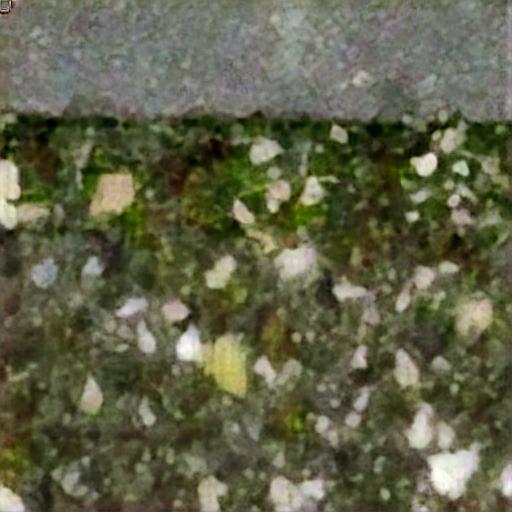} &
        \includegraphics[width=0.097\linewidth, height=0.097\linewidth]{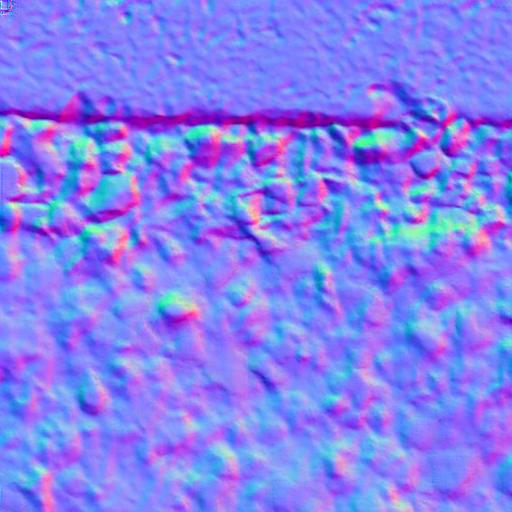} &
        \includegraphics[width=0.097\linewidth, height=0.097\linewidth]{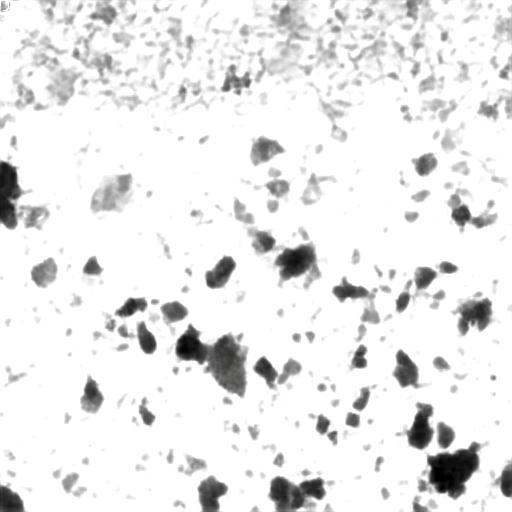} &
        \includegraphics[width=0.097\linewidth, height=0.097\linewidth]{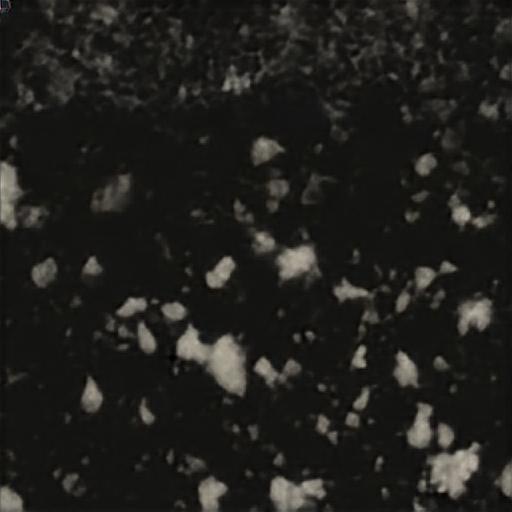} &
        \includegraphics[width=0.097\linewidth, height=0.097\linewidth]{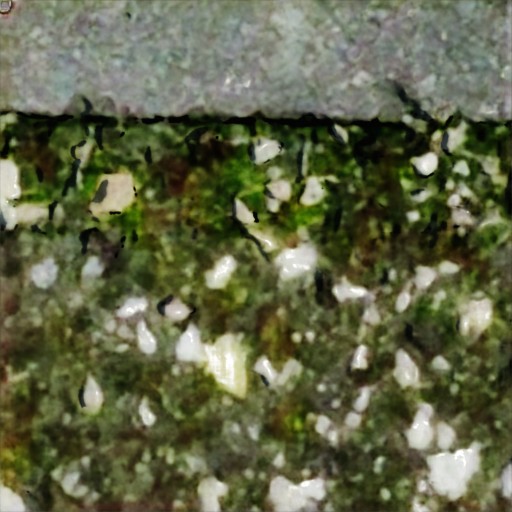} \\
    
        \vspace{-1mm}\hspace{-1mm}\includegraphics[width=0.097\linewidth, height=0.097\linewidth]{figures/edit/placeholder} &
        \includegraphics[width=0.097\linewidth, height=0.097\linewidth]{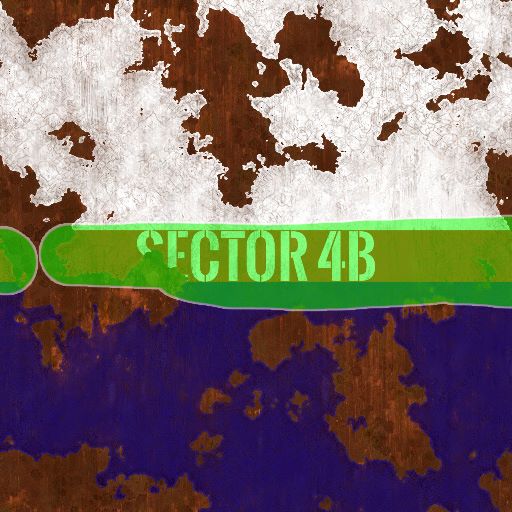} &
        \includegraphics[width=0.097\linewidth, height=0.097\linewidth]{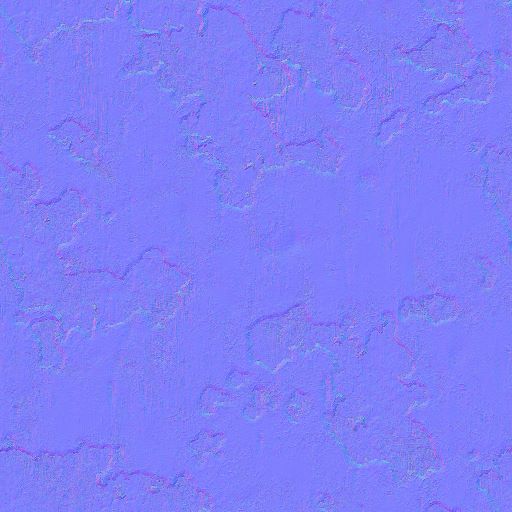} &
        \includegraphics[width=0.097\linewidth, height=0.097\linewidth]{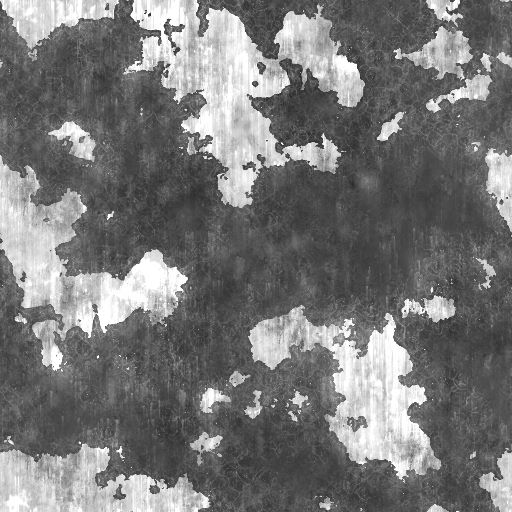} &
        \includegraphics[width=0.097\linewidth, height=0.097\linewidth]{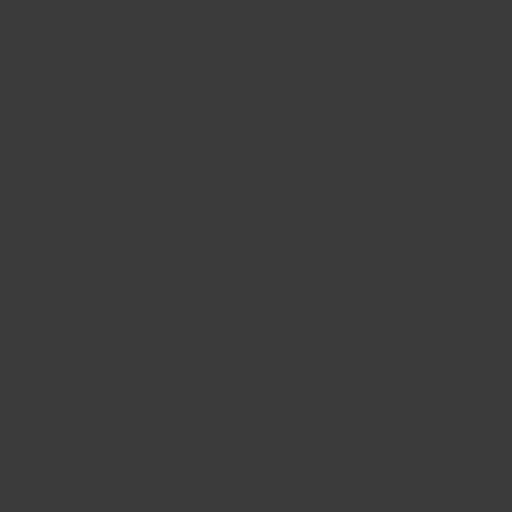} &
        \hspace{.5mm} \includegraphics[width=0.097\linewidth, height=0.097\linewidth]{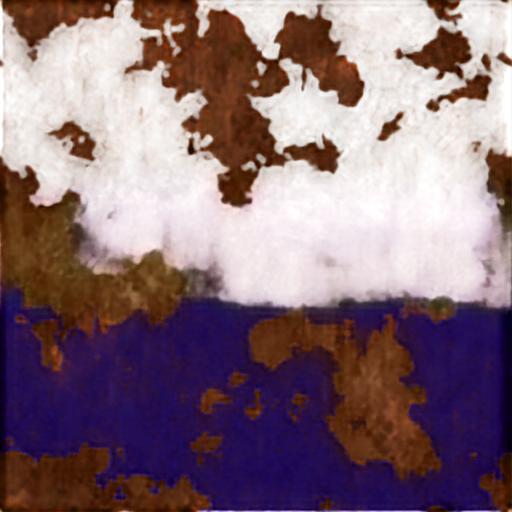} &
        \includegraphics[width=0.097\linewidth, height=0.097\linewidth]{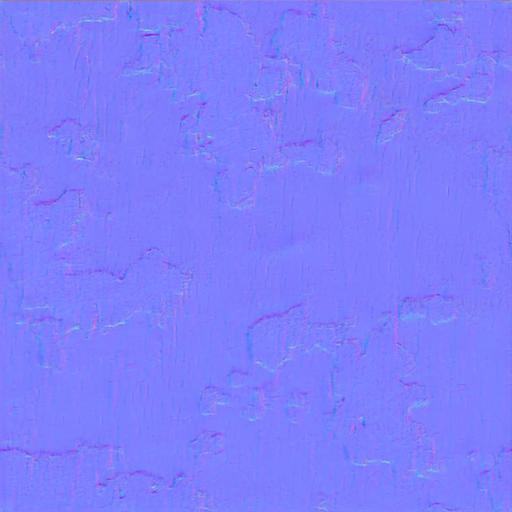} &
        \includegraphics[width=0.097\linewidth, height=0.097\linewidth]{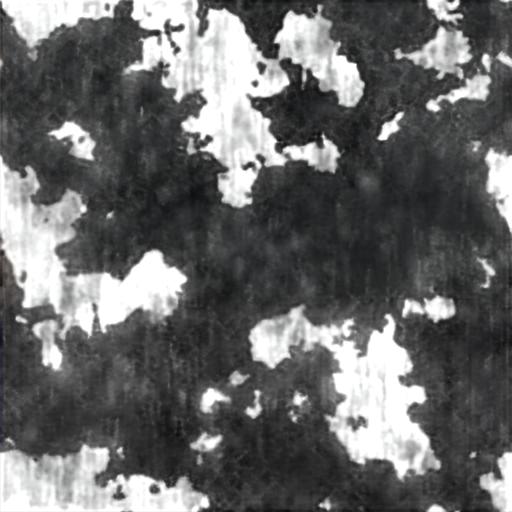} &
        \includegraphics[width=0.097\linewidth, height=0.097\linewidth]{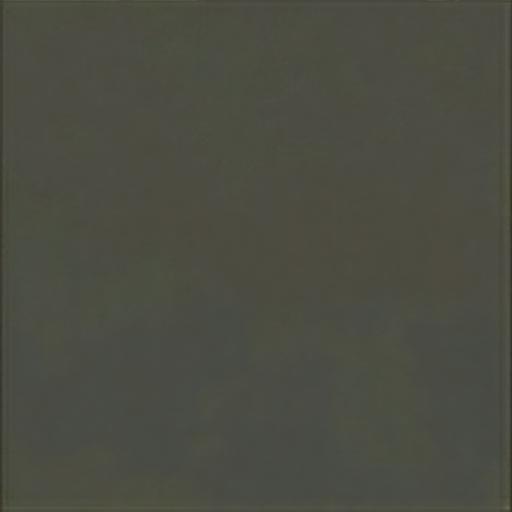} &
        \includegraphics[width=0.097\linewidth, height=0.097\linewidth]{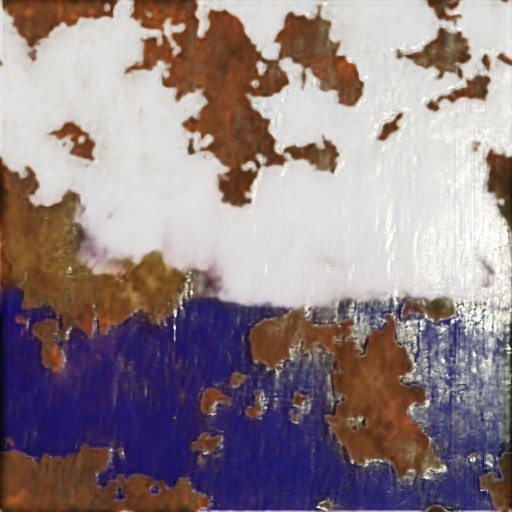} \\

        \vspace{-1mm}\hspace{-1mm}\makecell[b]{\footnotesize{``brick} \\ \footnotesize{wall''}\vspace{4.5mm}} &
        \includegraphics[width=0.097\linewidth, height=0.097\linewidth]{figures/edit/placeholder.jpg} &
        \includegraphics[width=0.097\linewidth, height=0.097\linewidth]{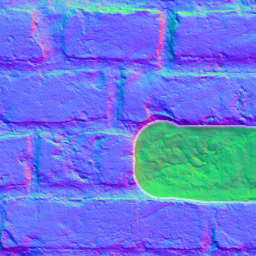} &
        \includegraphics[width=0.097\linewidth, height=0.097\linewidth]{figures/edit/placeholder.jpg} &
        \includegraphics[width=0.097\linewidth, height=0.097\linewidth]{figures/edit/placeholder.jpg} &
        \hspace{.5mm} \includegraphics[width=0.097\linewidth, height=0.097\linewidth]{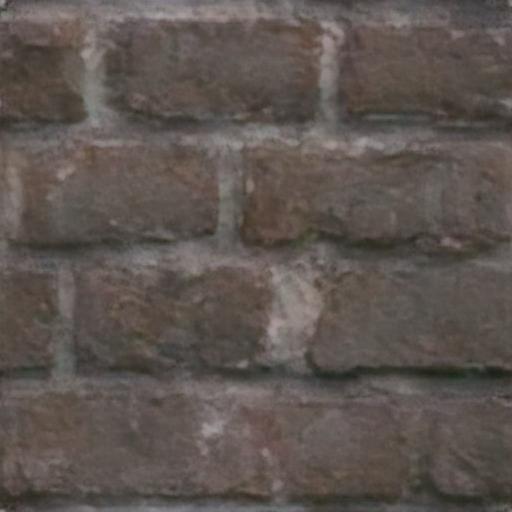} &
        \includegraphics[width=0.097\linewidth, height=0.097\linewidth]{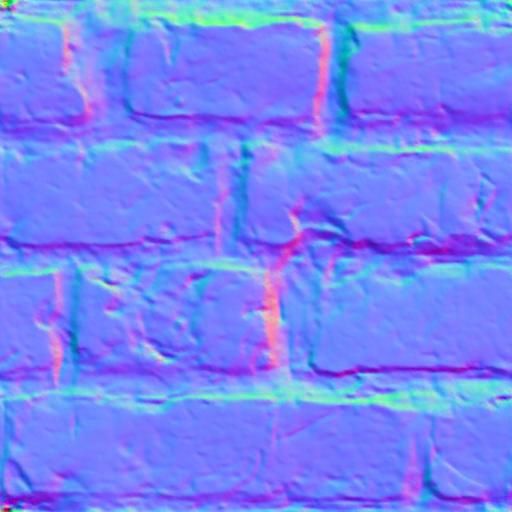} &
        \includegraphics[width=0.097\linewidth, height=0.097\linewidth]{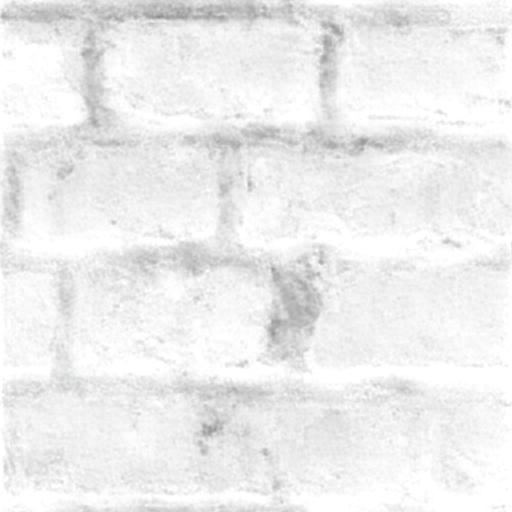} &
        \includegraphics[width=0.097\linewidth, height=0.097\linewidth]{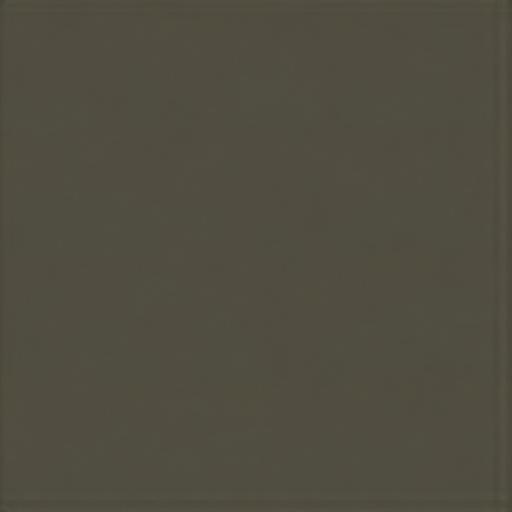} &
        \includegraphics[width=0.097\linewidth, height=0.097\linewidth]{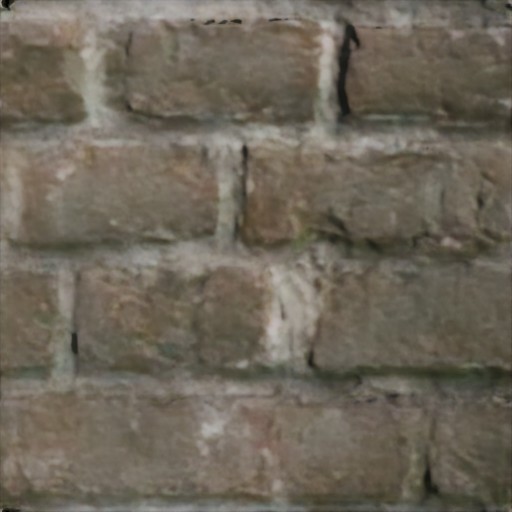} \\
    \end{tabular}
    
    \caption{\textbf{Material editing with inpainting}. The results show the flexibility of \modelname by being able to edit materials by inpainting, while using a condition (when provided) to determine the content of the generated part. The masked areas are highlighted in green, while fully masked maps are replaced with the `--' symbol. \suppmat{Additional samples are provided in the supplemental materials.}}
    \label{fig:edit_inpaint}
\end{figure*}

We demonstrate here the editing capabilities of \modelname which are made possible by the use of a multi-encoder architecture. In particular, the known structure in the latent space enables a deeper level of control over the generation by editing only specific maps or portions thereof through volumetric inpainting. Fig.~\ref{fig:edit_inpaint} shows the application of this technique to different use cases and its combination with multimodal conditioning.
Volumetric inpainting finds its most relevant application in generating missing maps for incomplete materials, as shown in the first and last rows of Fig.~\ref{fig:edit_inpaint}. Results show the method's ability to be coherent with the provided map structure (e.g., normal map in the last example) while capturing the semantics of the condition into the final material.

\subsection{Ablation Study}
\label{sec:ablation}

We perform an ablation study to substantiate our architectural design and training strategy choices evaluating the contribution of the multi-encoder architecture and of the rendering loss. 
\suppmat{Qualitative results for the ablation study are included in the supplemental materials.}

Results in Tab.~\ref{tab:ablation_arch} show the performance gain of the multi-encoder architecture compared to the baseline. This approach allows for better capturing map-specific features and efficiently compressing their information, resulting in a lower reconstruction distance between input and output.
To further substantiate our claim we explore different codebook size configurations for the baseline, ranging from 4096 codes to 16384 codes. 
Besides the lower reconstruction error, separate map representations give better control over the latent space and enable advanced material editing techniques.

We evaluate the contribution of the rendering loss $\mathcal{L}_\text{render}$~\cite{deschaintre2018single}, when training the VQ-GAN. Our baseline network is trained using the loss proposed in~\cite{rombach2022high}.
Results in Tab.~\ref{tab:ablation_loss} demonstrate that the introduction of the rendering loss improves the reconstruction quality by enforcing consistency in the rendered material.

\begin{table}[ht!]
    \begin{center}
        \begin{tabular}{lrrrrr}
        \toprule
        \textbf{Architecture} & \textbf{Diff.} & \textbf{Nrm.} & \textbf{Rgh.} & \textbf{Spec.} & \textbf{Rend.} \\
        \midrule
        Base \footnotesize(4096) & 0.057 & 0.061 & 0.114 & 0.166 & 0.267 \\
        Base \footnotesize(8192) & 0.049 & 0.052 & 0.098 & 0.144 & 0.233 \\
        Base \footnotesize(16384) & 0.047 & 0.051 & 0.102 & 0.152 & 0.227 \\
        Multi Enc. & \textbf{0.016} & \textbf{0.024} & \textbf{0.022} & \textbf{0.020} & \textbf{0.041} \\
        \bottomrule
        \end{tabular}
        \end{center}
        \caption{\textbf{Ablation study of architectural components}. Performance is measured in terms of RMSE between predicted and ground-truth maps. We report the codebook size between brackets for the \textit{``Base''} single encoder architecture.}
    \label{tab:ablation_arch}
\end{table}

\begin{table}[ht!]
\begin{center}
        \begin{tabular}{lrrrrr}
        \toprule
        \textbf{Losses} & \textbf{Diff.} & \textbf{Nrm.} & \textbf{Rgh.} & \textbf{Spec.} & \textbf{Rend.} \\
        \midrule
        $\sum\mathcal{L}$ from \cite{esser2021taming} & 0.038 & 0.030 & 0.047 & 0.033 & 0.064 \\
        \midrule
        \hspace{0.15cm}$+~\mathcal{L}_{\text{render}}$ & \textbf{0.016} & \textbf{0.024} & \textbf{0.022} & \textbf{0.020} & \textbf{0.041} \\
        
        \bottomrule
        \end{tabular}
        \end{center}
        \caption{\textbf{Ablation study of the contribution of the rendering loss when training the VQ-GAN}. Performance is measured in terms of RMSE between predicted and ground-truth maps.}
    \label{tab:ablation_loss}
\end{table}
\section{Limitations and Future Work}
\label{sec:limitation}
The generative capabilities of diffusion models come at the cost of computational resources. Although MatFuse is not explicitly limited to a specific resolution, 512$\times$512 generation takes \mytilde18 GB of GPU memory, with 768$\times$768 requiring slightly less than \mytilde24 GB.
Such memory consumption indeed limits the scalability of MatFuse to higher resolutions and, consequently, the representation of fine details, particularly for textures with high-frequency patterns. A potential solution could leverage a patch-based approach to alleviate computational burdens and enhance the model's applicability to higher resolutions.
Moreover, a noted limitation in the current implementation of MatFuse is the lack of tileability in the generated materials. This restricts the seamless use of synthesized textures for large surfaces. 
Additionally, it would be possible to use the generative capabilities of MatFuse to perform SVBRDF estimation from a single image by providing a picture as an additional local condition. This could be done in combination with a more advanced form of local conditioning such as ControlNet~\cite{zhang2023adding}.

\section{Conclusion}
\label{sec:conclusion}

In this paper, we present \modelname, a learning approach for the generation of materials in the form of reflectance maps with diffusion models.
The proposed approach specifically leverages the generative capabilities of recent diffusion methods to produce high-quality SVBRDF maps, supporting conditional and unconditional synthesis.

Inspired by the compositionality paradigm, \modelname supports extensive multimodal conditioning, thus providing control over the generation process. 
In particular, \modelname generates novel materials starting from a simple sketch, material samples, or textual descriptions, and supports conditioning combinations, e.g., sketch + color palette.
Additionally, \modelname introduces a novel ``volumetric inpainting'' strategy to perform map-level material editing. To do so, we propose a multi-encoder VQ-VAE, which learns a disentangled latent representation for each map.

\modelname can be a promising solution to build upon for material generation tasks, by extending the conditioning mechanism to include additional input modalities (e.g., semantic segmentation) and output controls (e.g., enforcing tileability), as well as exploring methodologies and architectures to support higher resolution with limited resources.

\section{Acknowledgments}
Research at the University of Catania is supported by the project Future Artificial Intelligence Research (FAIR) – PNRR MUR Cod. PE0000013 - CUP: E63C22001940006.
We thank Valentin Deschaintre for providing generation samples for the comparison with TileGen~\cite{zhou2022tilegen}.
{
    \small
    \bibliographystyle{ieeenat_fullname}
    \bibliography{main}

\begin{thebibliography}{54}
\providecommand{\natexlab}[1]{#1}
\providecommand{\url}[1]{\texttt{#1}}
\expandafter\ifx\csname urlstyle\endcsname\relax
  \providecommand{\doi}[1]{doi: #1}\else
  \providecommand{\doi}{doi: \begingroup \urlstyle{rm}\Url}\fi

\bibitem[Aittala et~al.(2016)Aittala, Aila, and
  Lehtinen]{aittala2016reflectance}
Miika Aittala, Timo Aila, and Jaakko Lehtinen.
\newblock Reflectance modeling by neural texture synthesis.
\newblock \emph{ACM Transactions on Graphics (ToG)}, 35\penalty0 (4):\penalty0
  1--13, 2016.

\bibitem[Arjovsky et~al.(2017)Arjovsky, Chintala, and
  Bottou]{arjovsky2017wasserstein}
Martin Arjovsky, Soumith Chintala, and L{\'e}on Bottou.
\newblock Wasserstein generative adversarial networks.
\newblock In \emph{International conference on machine learning}, pages
  214--223. PMLR, 2017.

\bibitem[Bi et~al.(2020)Bi, Xu, Sunkavalli, Kriegman, and
  Ramamoorthi]{bi2020deep}
Sai Bi, Zexiang Xu, Kalyan Sunkavalli, David Kriegman, and Ravi Ramamoorthi.
\newblock Deep 3d capture: Geometry and reflectance from sparse multi-view
  images.
\newblock In \emph{Proceedings of the IEEE/CVF Conference on Computer Vision
  and Pattern Recognition}, pages 5960--5969, 2020.

\bibitem[Brock et~al.(2018)Brock, Donahue, and Simonyan]{brock2018large}
Andrew Brock, Jeff Donahue, and Karen Simonyan.
\newblock Large scale gan training for high fidelity natural image synthesis.
\newblock \emph{arXiv preprint arXiv:1809.11096}, 2018.

\bibitem[Canny(1986)]{canny1986computational}
John Canny.
\newblock A computational approach to edge detection.
\newblock \emph{IEEE Transactions on Pattern Analysis and Machine
  Intelligence}, PAMI-8\penalty0 (6):\penalty0 679--698, 1986.

\bibitem[Deng et~al.(2009)Deng, Dong, Socher, Li, Li, and
  Fei-Fei]{deng2009imagenet}
Jia Deng, Wei Dong, Richard Socher, Li-Jia Li, Kai Li, and Li Fei-Fei.
\newblock Imagenet: A large-scale hierarchical image database.
\newblock In \emph{2009 IEEE Conference on Computer Vision and Pattern
  Recognition}, pages 248--255, 2009.

\bibitem[Deschaintre et~al.(2018)Deschaintre, Aittala, Durand, Drettakis, and
  Bousseau]{deschaintre2018single}
Valentin Deschaintre, Miika Aittala, Fredo Durand, George Drettakis, and Adrien
  Bousseau.
\newblock Single-image svbrdf capture with a rendering-aware deep network.
\newblock \emph{ACM Transactions on Graphics (ToG)}, 37\penalty0 (4):\penalty0
  1--15, 2018.

\bibitem[Deschaintre et~al.(2019)Deschaintre, Aittala, Durand, Drettakis, and
  Bousseau]{deschaintre2019flexible}
Valentin Deschaintre, Miika Aittala, Fr{\'e}do Durand, George Drettakis, and
  Adrien Bousseau.
\newblock Flexible {SVBRDF} capture with a multi-image deep network.
\newblock In \emph{Computer Graphics Forum}, pages 1--13. Wiley Online Library,
  2019.

\bibitem[Dhariwal and Nichol(2021)]{dhariwal2021diffusion}
Prafulla Dhariwal and Alexander Nichol.
\newblock Diffusion models beat gans on image synthesis.
\newblock \emph{Advances in Neural Information Processing Systems},
  34:\penalty0 8780--8794, 2021.

\bibitem[Dosovitskiy and Brox(2016)]{dosovitskiy2016generating}
Alexey Dosovitskiy and Thomas Brox.
\newblock Generating images with perceptual similarity metrics based on deep
  networks.
\newblock \emph{Advances in neural information processing systems}, 29, 2016.

\bibitem[Esser et~al.(2021)Esser, Rombach, and Ommer]{esser2021taming}
Patrick Esser, Robin Rombach, and Bjorn Ommer.
\newblock Taming transformers for high-resolution image synthesis.
\newblock In \emph{Proceedings of the IEEE/CVF conference on computer vision
  and pattern recognition}, pages 12873--12883, 2021.

\bibitem[Gao et~al.(2019)Gao, Li, Dong, Peers, Xu, and Tong]{gao2019deep}
Duan Gao, Xiao Li, Yue Dong, Pieter Peers, Kun Xu, and Xin Tong.
\newblock Deep inverse rendering for high-resolution svbrdf estimation from an
  arbitrary number of images.
\newblock \emph{ACM Trans. Graph.}, 38\penalty0 (4):\penalty0 134--1, 2019.

\bibitem[Goodfellow et~al.(2014)Goodfellow, Pouget-Abadie, Mirza, Xu,
  Warde-Farley, Ozair, Courville, and Bengio]{goodfellow2014generative}
Ian Goodfellow, Jean Pouget-Abadie, Mehdi Mirza, Bing Xu, David Warde-Farley,
  Sherjil Ozair, Aaron Courville, and Yoshua Bengio.
\newblock Generative adversarial nets.
\newblock In \emph{Advances in Neural Information Processing Systems}. Curran
  Associates, Inc., 2014.

\bibitem[Guarnera et~al.(2016)Guarnera, Guarnera, Ghosh, Denk, and
  Glencross]{guarnera2016brdf}
Darya Guarnera, Giuseppe~Claudio Guarnera, Abhijeet Ghosh, Cornelia Denk, and
  Mashhuda Glencross.
\newblock Brdf representation and acquisition.
\newblock In \emph{Computer Graphics Forum}, pages 625--650. Wiley Online
  Library, 2016.

\bibitem[Guehl et~al.(2020)Guehl, Allegre, Dischler, Benes, and
  Galin]{guehl2020semi}
Pascal Guehl, R{\'e}mi Allegre, J-M Dischler, Bedrich Benes, and Eric Galin.
\newblock Semi-procedural textures using point process texture basis functions.
\newblock In \emph{Computer Graphics Forum}, pages 159--171. Wiley Online
  Library, 2020.

\bibitem[Gulrajani et~al.(2017)Gulrajani, Ahmed, Arjovsky, Dumoulin, and
  Courville]{gulrajani2017improved}
Ishaan Gulrajani, Faruk Ahmed, Martin Arjovsky, Vincent Dumoulin, and Aaron~C
  Courville.
\newblock Improved training of wasserstein gans.
\newblock \emph{Advances in neural information processing systems}, 30, 2017.

\bibitem[Guo et~al.(2021)Guo, Lai, Tao, Cai, Wang, Guo, and
  Yan]{guo2021highlight}
Jie Guo, Shuichang Lai, Chengzhi Tao, Yuelong Cai, Lei Wang, Yanwen Guo, and
  Ling-Qi Yan.
\newblock Highlight-aware two-stream network for single-image svbrdf
  acquisition.
\newblock \emph{ACM Transactions on Graphics (TOG)}, 40\penalty0 (4):\penalty0
  1--14, 2021.

\bibitem[Guo et~al.(2020)Guo, Smith, Ha{\v{s}}an, Sunkavalli, and
  Zhao]{guo2020materialgan}
Yu Guo, Cameron Smith, Milo{\v{s}} Ha{\v{s}}an, Kalyan Sunkavalli, and Shuang
  Zhao.
\newblock Material{GAN}: reflectance capture using a generative svbrdf model.
\newblock \emph{arXiv preprint arXiv:2010.00114}, 2020.

\bibitem[He et~al.(2023)He, Guo, Zhang, Tu, Chen, Guo, Wang, and
  Dai]{guo2023text2mat}
Zhen He, Jie Guo, Yan Zhang, Qinghao Tu, Mufan Chen, Yanwen Guo, Pengyu Wang,
  and Wei Dai.
\newblock {Text2Mat: Generating Materials from Text}.
\newblock In \emph{Pacific Graphics Short Papers and Posters}. The Eurographics
  Association, 2023.

\bibitem[Heusel et~al.(2017)Heusel, Ramsauer, Unterthiner, Nessler, and
  Hochreiter]{heusel2017gans}
Martin Heusel, Hubert Ramsauer, Thomas Unterthiner, Bernhard Nessler, and Sepp
  Hochreiter.
\newblock Gans trained by a two time-scale update rule converge to a local nash
  equilibrium.
\newblock In \emph{Advances in Neural Information Processing Systems}, 2017.

\bibitem[Ho and Salimans(2022)]{ho2022classifier}
Jonathan Ho and Tim Salimans.
\newblock Classifier-free diffusion guidance.
\newblock \emph{arXiv preprint arXiv:2207.12598}, 2022.

\bibitem[Ho et~al.(2020)Ho, Jain, and Abbeel]{ho2020denoising}
Jonathan Ho, Ajay Jain, and Pieter Abbeel.
\newblock Denoising diffusion probabilistic models.
\newblock \emph{Advances in Neural Information Processing Systems},
  33:\penalty0 6840--6851, 2020.

\bibitem[Hu et~al.(2022)Hu, Ha{\v{s}}an, Guerrero, Rushmeier, and
  Deschaintre]{hu2022controlling}
Yiwei Hu, Milo{\v{s}} Ha{\v{s}}an, Paul Guerrero, Holly Rushmeier, and Valentin
  Deschaintre.
\newblock Controlling material appearance by examples.
\newblock In \emph{Computer Graphics Forum}, pages 117--128. Wiley Online
  Library, 2022.

\bibitem[Huang et~al.(2023)Huang, Chen, Liu, Shen, Zhao, and
  Zhou]{huang2023composer}
Lianghua Huang, Di Chen, Yu Liu, Yujun Shen, Deli Zhao, and Jingren Zhou.
\newblock Composer: Creative and controllable image synthesis with composable
  conditions.
\newblock \emph{arXiv preprint arXiv:2302.09778}, 2023.

\bibitem[Isola et~al.(2017)Isola, Zhu, Zhou, and Efros]{isola2017image}
Phillip Isola, Jun-Yan Zhu, Tinghui Zhou, and Alexei~A Efros.
\newblock Image-to-image translation with conditional adversarial networks.
\newblock In \emph{Proceedings of the IEEE conference on computer vision and
  pattern recognition}, pages 1125--1134, 2017.

\bibitem[Karras et~al.(2017)Karras, Aila, Laine, and
  Lehtinen]{karras2017progressive}
Tero Karras, Timo Aila, Samuli Laine, and Jaakko Lehtinen.
\newblock Progressive growing of gans for improved quality, stability, and
  variation.
\newblock \emph{arXiv preprint arXiv:1710.10196}, 2017.

\bibitem[Karras et~al.(2020{\natexlab{a}})Karras, Laine, Aittala, Hellsten,
  Lehtinen, and Aila]{karras2020analyzing}
Tero Karras, Samuli Laine, Miika Aittala, Janne Hellsten, Jaakko Lehtinen, and
  Timo Aila.
\newblock Analyzing and improving the image quality of stylegan.
\newblock In \emph{Proceedings of the IEEE/CVF conference on computer vision
  and pattern recognition}, pages 8110--8119, 2020{\natexlab{a}}.

\bibitem[Karras et~al.(2020{\natexlab{b}})Karras, Laine, Aittala, Hellsten,
  Lehtinen, and Aila]{stylegan2}
Tero Karras, Samuli Laine, Miika Aittala, Janne Hellsten, Jaakko Lehtinen, and
  Timo Aila.
\newblock Analyzing and improving the image quality of stylegan.
\newblock In \emph{Proceedings of the IEEE/CVF conference on computer vision
  and pattern recognition}, pages 8110--8119, 2020{\natexlab{b}}.

\bibitem[Kingma and Ba(2014)]{adam}
Diederik~P Kingma and Jimmy Ba.
\newblock Adam: A method for stochastic optimization.
\newblock \emph{arXiv preprint arXiv:1412.6980}, 2014.

\bibitem[Lake et~al.(2017)Lake, Ullman, Tenenbaum, and
  Gershman]{lake2017building}
Brenden~M Lake, Tomer~D Ullman, Joshua~B Tenenbaum, and Samuel~J Gershman.
\newblock Building machines that learn and think like people.
\newblock \emph{Behavioral and brain sciences}, 40:\penalty0 e253, 2017.

\bibitem[Li et~al.(2017)Li, Dong, Peers, and Tong]{li2017modeling}
Xiao Li, Yue Dong, Pieter Peers, and Xin Tong.
\newblock Modeling surface appearance from a single photograph using
  self-augmented convolutional neural networks.
\newblock \emph{ACM Transactions on Graphics (ToG)}, 36\penalty0 (4):\penalty0
  1--11, 2017.

\bibitem[Li et~al.(2018)Li, Sunkavalli, and Chandraker]{li2018materials}
Zhengqin Li, Kalyan Sunkavalli, and Manmohan Chandraker.
\newblock Materials for masses: Svbrdf acquisition with a single mobile phone
  image.
\newblock In \emph{Proceedings of the European Conference on Computer Vision
  (ECCV)}, pages 72--87, 2018.

\bibitem[Loshchilov and Hutter(2017)]{adamw}
Ilya Loshchilov and Frank Hutter.
\newblock Decoupled weight decay regularization.
\newblock \emph{arXiv preprint arXiv:1711.05101}, 2017.

\bibitem[Martin et~al.(2022)Martin, Roullier, Rouffet, Kaiser, and
  Boubekeur]{martin2022materia}
Rosalie Martin, Arthur Roullier, Romain Rouffet, Adrien Kaiser, and Tamy
  Boubekeur.
\newblock Materia: Single image high-resolution material capture in the wild.
\newblock In \emph{Computer Graphics Forum}, pages 163--177. Wiley Online
  Library, 2022.

\bibitem[Mescheder(2018)]{mescheder2018convergence}
Lars Mescheder.
\newblock On the convergence properties of gan training.
\newblock \emph{arXiv preprint arXiv:1801.04406}, 1:\penalty0 16, 2018.

\bibitem[Metz et~al.(2016)Metz, Poole, Pfau, and
  Sohl-Dickstein]{metz2016unrolled}
Luke Metz, Ben Poole, David Pfau, and Jascha Sohl-Dickstein.
\newblock Unrolled generative adversarial networks.
\newblock \emph{arXiv preprint arXiv:1611.02163}, 2016.

\bibitem[Radford et~al.(2021)Radford, Kim, Hallacy, Ramesh, Goh, Agarwal,
  Sastry, Askell, Mishkin, Clark, et~al.]{clip}
Alec Radford, Jong~Wook Kim, Chris Hallacy, Aditya Ramesh, Gabriel Goh,
  Sandhini Agarwal, Girish Sastry, Amanda Askell, Pamela Mishkin, Jack Clark,
  et~al.
\newblock Learning transferable visual models from natural language
  supervision.
\newblock In \emph{International conference on machine learning}, pages
  8748--8763. PMLR, 2021.

\bibitem[Raistrick et~al.(2023)Raistrick, Lipson, Ma, Mei, Wang, Zuo, Kayan,
  Wen, Han, Wang, Newell, Law, Goyal, Yang, and Deng]{infinigen2023infinite}
Alexander Raistrick, Lahav Lipson, Zeyu Ma, Lingjie Mei, Mingzhe Wang, Yiming
  Zuo, Karhan Kayan, Hongyu Wen, Beining Han, Yihan Wang, Alejandro Newell, Hei
  Law, Ankit Goyal, Kaiyu Yang, and Jia Deng.
\newblock Infinite photorealistic worlds using procedural generation.
\newblock In \emph{Proceedings of the IEEE/CVF Conference on Computer Vision
  and Pattern Recognition}, pages 12630--12641, 2023.

\bibitem[Rombach et~al.(2022)Rombach, Blattmann, Lorenz, Esser, and
  Ommer]{rombach2022high}
Robin Rombach, Andreas Blattmann, Dominik Lorenz, Patrick Esser, and Bj{\"o}rn
  Ommer.
\newblock High-resolution image synthesis with latent diffusion models.
\newblock In \emph{Proceedings of the IEEE/CVF Conference on Computer Vision
  and Pattern Recognition}, pages 10684--10695, 2022.

\bibitem[Ronneberger et~al.(2015)Ronneberger, Fischer, and
  Brox]{ronneberger2015u}
Olaf Ronneberger, Philipp Fischer, and Thomas Brox.
\newblock U-net: Convolutional networks for biomedical image segmentation.
\newblock In \emph{International Conference on Medical image computing and
  computer-assisted intervention}, pages 234--241. Springer, 2015.

\bibitem[Salimans et~al.(2016)Salimans, Goodfellow, Zaremba, Cheung, Radford,
  and Chen]{salimans2016improved}
Tim Salimans, Ian Goodfellow, Wojciech Zaremba, Vicki Cheung, Alec Radford, and
  Xi Chen.
\newblock Improved techniques for training gans.
\newblock \emph{Advances in neural information processing systems}, 29, 2016.

\bibitem[Sohl-Dickstein et~al.(2015)Sohl-Dickstein, Weiss, Maheswaranathan, and
  Ganguli]{sohl2015deep}
Jascha Sohl-Dickstein, Eric Weiss, Niru Maheswaranathan, and Surya Ganguli.
\newblock Deep unsupervised learning using nonequilibrium thermodynamics.
\newblock In \emph{International Conference on Machine Learning}, pages
  2256--2265. PMLR, 2015.

\bibitem[Song et~al.(2020)Song, Meng, and Ermon]{song2020denoising}
Jiaming Song, Chenlin Meng, and Stefano Ermon.
\newblock Denoising diffusion implicit models.
\newblock \emph{arXiv preprint arXiv:2010.02502}, 2020.

\bibitem[Song et~al.(2023)Song, He, Li, Ma, Ming, Mao, Pei, Peng, Hu, Yao, and
  Zhang]{song2023synthetic}
Zhihang Song, Zimin He, Xingyu Li, Qiming Ma, Ruibo Ming, Zhiqi Mao, Huaxin
  Pei, Lihui Peng, Jianming Hu, Danya Yao, and Yi Zhang.
\newblock Synthetic datasets for autonomous driving: A survey, 2023.

\bibitem[Szegedy et~al.(2014)Szegedy, Liu, Jia, Sermanet, Reed, Anguelov,
  Erhan, Vanhoucke, and Rabinovich]{szegedy2014going}
Christian Szegedy, Wei Liu, Yangqing Jia, Pierre Sermanet, Scott~E Reed,
  Dragomir Anguelov, Dumitru Erhan, Vincent Vanhoucke, and Andrew Rabinovich.
\newblock Going deeper with convolutions. corr abs/1409.4842 (2014), 2014.

\bibitem[Van Den~Oord et~al.(2017)Van Den~Oord, Vinyals, et~al.]{van2017neural}
Aaron Van Den~Oord, Oriol Vinyals, et~al.
\newblock Neural discrete representation learning.
\newblock \emph{Advances in neural information processing systems}, 30, 2017.

\bibitem[Vaswani et~al.(2017)Vaswani, Shazeer, Parmar, Uszkoreit, Jones, Gomez,
  Kaiser, and Polosukhin]{vaswani2017attention}
Ashish Vaswani, Noam Shazeer, Niki Parmar, Jakob Uszkoreit, Llion Jones,
  Aidan~N Gomez, {\L}ukasz Kaiser, and Illia Polosukhin.
\newblock Attention is all you need.
\newblock \emph{Advances in neural information processing systems}, 30, 2017.

\bibitem[Vecchio et~al.(2021)Vecchio, Palazzo, and
  Spampinato]{vecchio2021surfacenet}
Giuseppe Vecchio, Simone Palazzo, and Concetto Spampinato.
\newblock Surfacenet: Adversarial svbrdf estimation from a single image.
\newblock In \emph{Proceedings of the IEEE/CVF International Conference on
  Computer Vision}, pages 12840--12848, 2021.

\bibitem[Vecchio et~al.(2022)Vecchio, Palazzo, Guastella, Carlucho, Albrecht,
  Muscato, and Spampinato]{vecchio2022midgard}
Giuseppe Vecchio, Simone Palazzo, Dario~C Guastella, Ignacio Carlucho,
  Stefano~V Albrecht, Giovanni Muscato, and Concetto Spampinato.
\newblock Midgard: A simulation platform for autonomous navigation in
  unstructured environments.
\newblock \emph{arXiv preprint arXiv:2205.08389}, 2022.

\bibitem[Wang et~al.(2023)Wang, Chan, and Loy]{wang2023exploring}
Jianyi Wang, Kelvin~CK Chan, and Chen~Change Loy.
\newblock Exploring clip for assessing the look and feel of images.
\newblock In \emph{Proceedings of the AAAI Conference on Artificial
  Intelligence}, pages 2555--2563, 2023.

\bibitem[Zhang et~al.(2021)Zhang, Liang, Van~Gool, and
  Timofte]{zhang2021designing}
Kai Zhang, Jingyun Liang, Luc Van~Gool, and Radu Timofte.
\newblock Designing a practical degradation model for deep blind image
  super-resolution.
\newblock In \emph{Proceedings of the IEEE/CVF International Conference on
  Computer Vision}, pages 4791--4800, 2021.

\bibitem[Zhang and Agrawala(2023)]{zhang2023adding}
Lvmin Zhang and Maneesh Agrawala.
\newblock Adding conditional control to text-to-image diffusion models.
\newblock \emph{arXiv preprint arXiv:2302.05543}, 2023.

\bibitem[Zhou and Kalantari(2021)]{zhou2021adversarial}
Xilong Zhou and Nima~Khademi Kalantari.
\newblock Adversarial single-image svbrdf estimation with hybrid training.
\newblock In \emph{Computer Graphics Forum}, pages 315--325. Wiley Online
  Library, 2021.

\bibitem[Zhou et~al.(2022)Zhou, Hasan, Deschaintre, Guerrero, Sunkavalli, and
  Kalantari]{zhou2022tilegen}
Xilong Zhou, Milos Hasan, Valentin Deschaintre, Paul Guerrero, Kalyan
  Sunkavalli, and Nima~Khademi Kalantari.
\newblock Tile{Gen}: Tileable, controllable material generation and capture.
\newblock In \emph{SIGGRAPH Asia 2022 Conference Papers}, pages 1--9, 2022.

\end{thebibliography}
}

\end{document}